\documentclass[letterpaper]{article} % DO NOT CHANGE THIS
\usepackage[]{aaai2026}  % DO NOT CHANGE THIS
\usepackage{times}  % DO NOT CHANGE THIS
\usepackage{helvet}  % DO NOT CHANGE THIS
\usepackage{courier}  % DO NOT CHANGE THIS
\usepackage[hyphens]{url}  % DO NOT CHANGE THIS
\usepackage{graphicx} % DO NOT CHANGE THIS
\usepackage{natbib}  % DO NOT CHANGE THIS AND DO NOT ADD ANY OPTIONS TO IT

\usepackage{booktabs}
\usepackage{subcaption}

\usepackage{caption} % DO NOT CHANGE THIS AND DO NOT ADD ANY OPTIONS TO IT
\usepackage{algorithm}
\usepackage{algorithmic}
\usepackage{xcolor}

\usepackage{multirow}
\usepackage{newfloat}
\usepackage{listings}
\DeclareCaptionStyle{ruled}{labelfont=normalfont,labelsep=colon,strut=off} % DO NOT CHANGE THIS
\floatstyle{ruled}
\newfloat{listing}{tb}{lst}{}
\floatname{listing}{Listing}
\title{On Identifying Why and When  Foundation Models  Perform Well on Time-Series Forecasting Using Automated Explanations and Rating}
\author{
Michael "Xander" Widener,
Kausik Lakkaraju,
John Aydin,
Biplav Srivastava
}
\affiliations{
   Department of Computer Science and Engineering, University of South Carolina, Columbia, SC, USA
}

\usepackage{amsmath}
\usepackage{bibentry}
\begin{document}

\maketitle
\begin{abstract}
Time-series forecasting models (TSFM) have evolved from classical statistical methods to sophisticated foundation models, yet understanding why and when these models succeed or fail remains challenging. Despite this known limitation, time series forecasting models are increasingly used to generate information that informs real-world actions with equally real consequences. Understanding the complexity, performance variability, and opaque nature of these models then becomes a valuable endeavor to combat serious concerns about how users should interact with and rely on these models' outputs.

This work addresses these concerns by combining traditional explainable AI (XAI) methods with Rating Driven Explanations (RDE) to assess TSFM performance and interpretability across diverse domains and use cases. We evaluate four distinct model architectures: ARIMA, Gradient Boosting, Chronos (time-series specific foundation model), Llama (general-purpose; both fine-tuned and base models) on four heterogeneous datasets spanning finance, energy, transportation, and automotive sales domains. In doing so, we demonstrate that feature-engineered models (e.g., Gradient Boosting) consistently outperform foundation models (e.g., Chronos) in volatile or sparse domains (e.g., power, car parts) while providing more interpretable explanations, whereas foundation models excel only in stable or trend-driven contexts (e.g., finance). 
%We further show that \xander{XAI \& RDE insights}

\end{abstract}

\section{Introduction}
\label{sec:introduction}
Time-series forecasting underpins critical decisions in domains such as finance, energy management, and logistics, yet model performance can vary widely across domains. Classical statistical methods like ARIMA \cite{shumway2017arima} are interpretable, but often underperform on volatile or sparse data \cite{hybrid-svd}. Modern approaches, ranging from feature-engineered gradient boosting \cite{agapitos2017regularised} to foundation models like Chronos \cite{ansari2024chronos}, deliver high predictive accuracy but introduce opacity, raising concerns about trust and reliability. Prior work \cite{classical-vs-dl-llm-ts} shows the feature engineered gradient boosting can outperform all other evaluated models, including naive gradient boosting and ARIMA (an observation we also confirm in our experiments).

EXplainable AI (XAI) methods aim to clarify model behaviour by identifying how input features contribute to predictions \cite{arrieta2020explainable}. Techniques such as SHAP \cite{lundberg2017shap} and LIME \cite{ribeiro2016lime} attribute importance at the individual prediction level (local explanations), while global approaches such as global SHAP and TsSHAP \cite{raykar2023tsshap} provide feature attribution across the dataset. These methods are effective for showing why a model made a given prediction, but they do not indicate whether the model behaves fairly across subgroups \cite{alikhademi2021can} or remains reliable under different input conditions. To address this limitation, we use Rating-Driven Explanations (RDE) \cite{lakkaraju2025h-xai}, a causally grounded approach derived from the rating method \cite{kausik2024rating,lakkaraju2025creating}. RDE measures how model outcomes change under interventions, quantifies robustness, which includes models' sensitivity to confounders and protected attributes, and compares results against biased and random baselines. This approach provides explanation from a model stability and fairness from a fairness perspective.

In this work, we use SHAP \cite{lundberg2017shap} and LIME \cite{ribeiro2016lime} for providing local explanations, global SHAP and TsSHAP \cite{raykar2023tsshap} for global level feature attributions, and RDE for explanations about robustness of the models. Specifically, we address three key research questions: 

\noindent\textbf{RQ1:} When forecasting models succeed or
fail?

\noindent\textbf{RQ2:} What do common XAI methods tell us
about why forecasting models succeed or fail?

\noindent\textbf{RQ3:} How can rating augment our understanding of when and why forecasting models succeed or fail?

\begin{table}
    \tiny
    \centering
    \begin{tabular}{c|c|c|c|c}
        \toprule
            \textbf{Dataset} & \textbf{Domain} & \textbf{Number of Series} & \textbf{Min Length} & \textbf{Max Length} \\
        \midrule
        \textbf{Finance}    & Finance   & $6$    & $250$   & $250$ \\
        \textbf{Power}      & Energy    & $1$    & $20,915$ & $20,915$ \\
        \textbf{Pedestrian} & Mobility  & $66$   & $576$   & $96,424$ \\
        \textbf{Car Parts}  & Sales     & $2,674$ & $51$    & $51$ \\
        \bottomrule
    \end{tabular}
    \caption{  We selected the pedestrian and Car Parts datasets from Monash \cite{godahewa2021monashtimeseriesforecasting}, power data from collected by Tantiv4 sensors \cite{muppasani2023dataset}, and Yahoo Finance as representative open time-series datasets spanning diverse domains. Other widely used benchmarks (e.g., M4, ETT, WeatherBench) were excluded to maintain tractability, though inclusion in future work would further strengthen generalizability claims.}
    \label{tab:datasets_used}
  
\end{table}

To explore these questions, we conduct experiments on four benchmark datasets (see Table~\ref{tab:datasets_used}), which vary in frequency, volatility, and periodicity. We compare a range of forecasting models spanning classical, machine learning, and foundation-model approaches as seen in Table~\ref{tab:model_performance}.
\begin{table}
    \tiny
    \centering
    \begin{tabular}{l|c|c|c|c}
        \toprule
        \textbf{Model} & \textbf{Category} & \textbf{Params} & \textbf{Pretrained?} & \textbf{Needs FE?} \\
        \midrule
        \textbf{ARIMA} & Statistical & -- & No & No \\
        \textbf{Gradient Boosting} & ML (Tabular) & \textasciitilde100k–300k & No & Yes \\
        \textbf{Chronos} & Foundation Model &  47.7M& Yes & No \\
        \textbf{LLaMA (Finetuned)} & LLM-based & \textasciitilde7B+ & Yes & No \\
        \bottomrule
    \end{tabular}
    \caption{Models used. parameters size $\rightarrow$ "Params" and reflects typical scale; feature engineering $\rightarrow$  "FE".}
    \label{tab:models_used}
\end{table}

% \biplav{No mention of explanation methods.}
Our results reveal that feature-engineered models outperform foundation models in volatile domains (e.g., energy) while offering better interpretability, whereas foundation models excel in stable, trend-driven contexts (e.g., finance). These findings provide actionable guidance for selecting forecasting approaches based on domain requirements.

% --------------------
% --------------------
\section{Related Work}
In this section, we situate our work within relevant prior work.
\subsection{Time-Series Forecasting Models (TSFM)}

Foundation models (FMs), including large language models (LLMs), have recently been adapted for time-series forecasting \cite{zhou2023one,jin2023time,pan2024s}.  Fine-tuned language-pretrained transformers \cite{zhou2023one,jin2023time} and parameter-efficient adaptations \cite{gruver2024large,cao2023tempo,ekambaram2024tiny} show competitive performance across diverse tasks. Chronos \cite{ansari2024chronos} and Unified-TS \cite{woo2024unified} improve accuracy and generalization, while \cite{rasul2023lag,das2023decoder} explore new tokenization schemes. In explainable forecasting, \cite{yu2023temporal} integrates historical prices, metadata, and news with LLMs to produce predictions and rationales. Lightweight designs \cite{garza2023timegpt,ekambaram2024tiny} enable real-time use, and multi-pattern integration \cite{talukder2024totem} improves precision. Models trained from scratch, such as Lag-Llama \cite{gruver2024large}, achieve strong zero-shot performance, with further gains from temporal decomposition \cite{cao2023tempo} and hybrid encoder–decoder designs \cite{goswami2024moment}.

Fine-tuning remains important even with retrieval-based methods. RAF \cite{tire2024retrieval} and FinSrag \cite{xiao2025retrieval} both report higher accuracy from domain-specific adaptation. Chronos \cite{ansari2024chronos} discretizes time-series through scaling and quantization, while Chronos-Bolt uses a patch-based representation and decoder outputs to improve both forecasting accuracy and inference speed. In this work, we employ Chronos-Bolt alongside a widely used general-purpose foundation model, LLaMA, and its fine-tuned variants as TSFM.

\subsection{Explainability in AI}

Explainable AI (XAI) refers to processes and methods that help describe how AI models work \cite{arrieta2020explainable}. Feature importance methods such as SHapley Additive exPlanations (SHAP) and counterfactual explanations are the most widely-used explanation methods. The former is sensitive to feature correlations and the latter only explains a single outcome instead of the entire model \cite{alfeo2023local}.
Users benefit more from integrated local and global explanations, yet the majority of XAI methods have focused on the former \cite{hoffman2023explainable}.  We use traditional XAI methods, to explain predictions, both at the local and global level, but these do not say much about a model's overall fairness or robustness \cite{alikhademi2021can}. For that, we turn to RDE, which is causally grounded and better suited to communicate how the model behaves across different scenarios on a global level.

\subsection{Explainability in Time-Series Forecasting Models (TSFMs)}
The most common methods in time-series forecasting \cite{arsenault2025survey} are Local Interpretable Model-agnostic Explanations (LIME) \cite{ribeiro2016lime} \& SHapley Additive exPlanations (SHAP) \cite{lundberg2017shap}. SHAP enables users to get a sense of interpretable models' feature attribution with both magnitude and direction. Similar to SHAP \cite{lundberg2017shap}, LIME \cite{ribeiro2016lime} calculates feature attributions using a local linear model to estimate the importance of attributes in a task. When adapted to time-series forecasting, it can be used to examine feature importance within a model or gain insights into the influence of different segments of a time-series on the forecast \cite{schlegel2021tsmulelocalinterpretablemodelagnostic}.

Traditional XAI methods for TSFMs rely on feature attribution and visualization methods. Model-agnostic methods such as SHAP are widely used~\cite{mokhtari2019interpreting, schlegel2019towards, zhang2022fi}, along with partial dependence plots \cite{greenwell2017pdp} to estimate marginal contributions of input variables and forecast outputs. These methods have been adapted to consider time-dependent inputs, but they are often limited in capturing causality or robustness under input perturbation \cite{lakkaraju2025creating}. In deep learning-based TSFMs, attention mechanisms have also been used to show influential time steps or variables in multivariate forecasting tasks \cite{abbasimehr2022improving, li2019ea, shih2019temporal}. However, such mechanisms are typically embedded within the models and therefore may not be reliable indicators to measure causal influence. Though these methods have been useful for prediction interpretability, they tend to disproportionately focus on either local or global explanations. SHAP often provides local insights into single predictions. PDPs, for instance, offer aggregated trends across a dataset. In TSFMs, it is often necessary to combine both perspectives, i.e., understanding how individual time points contribute to certain forecasts (local), and how these patterns also generalize across entire data (global). Existing approaches do not always consistently support both forms of explanation in an integrated or interactive manner, especially under stakeholder-specific constraints.
In recent years, there has been a growing shift in the literature toward methods that not only explain model predictions but also empower stakeholders at different levels to interrogate model behavior through hypothesis testing and counterfactual reasoning~\cite{ccelik2023extending, tasin2023diabetes}.

While these methods help magnify the explainability of already interpretable models (GB) some significant concessions are often required to explain more complex black box models (segment-based LIME, TsSHAP, ...).
%\xander{John, write about the limitations found in the literature for LIME}  
Adapting LIME to time-series forecasting is a non-trivial task. \cite{limesegment} propose an adaptation for classification, implementing LIME for time-series with a novel segmentation algorithm, a perturbation technique which aims to mimic realistic background noise, and distance algorithm specifically designed to compare time-series. \cite{schlegel2021tsmulelocalinterpretablemodelagnostic} adapt LIME to forecasting, testing several segmentation algorithms and simple perturbation strategies. For the sake of simplicity, we choose to use their work to perform our LIME analysis, focusing on the baseline segmentation algorithm, a uniform segmentation of the time-series, with perturbation strategies chosen for each model. 
\vspace{-0.2em}
\subsection{Rating of AI Models}
\vspace{-1em}
Prior work has developed rating methods to assess AI models from a third-party perspective, often without access to training data. Early work \cite{srivastava2020rating} measured bias, such as gender bias in machine translation \cite{srivastava2018towards}, and used visualizations to communicate results \cite{vega-tool,vega-userstudy-translatorbias}, but lacked a causal foundation. More recent approaches introduced causal rating methods to isolate the effect of protected attributes on model outputs. These were applied first to sentiment analysis \cite{kausik2024rating}, later to composite NLP tasks \cite{kausik2023the}, and then to time-series forecasting \cite{lakkaraju2024timeseries,lakkaraju2025creating}. In this work, we build on that foundation, using ratings not only as performance metrics, but also as Rating-Driven Explanations (RDE), originally introduced in \cite{lakkaraju2025h-xai}. This allows us to describe how models behave when sensitive attributes vary, complementing traditional XAI methods that focus on individual predictions or global feature importance.
\section{Problem Setting}
\label{sec:problem-setting}
\subsection{Mathematical Formulation}
\vspace{-1em}
Let $\mathcal{D} = {D_1, D_2, D_3, D_4}$ denote our collection of four datasets from distinct domains, each described in detail in Section \ref{sec:data}. For each dataset $D_i$, we observe a univariate time-series $\mathbf{x}^{(i)} = \{x_1^{(i)}, x_2^{(i)}, \ldots, x_{T_i}^{(i)}\}$ where $T_i$ represents the total length of series $i$. While three datasets consist of continuous-time observations, the financial dataset comprises business daily observations (excluding weekends and holidays), which we treat as continuous for the purpose of forecasting.

\subsection{Forecasting Task}

Given a time-series $\mathbf{x}^{(i)}$ of length $T_i$, we partition the data into training and test sets using an 80-20 split:
\begin{align}
\mathbf{x}_{\text{train}}^{(i)} &= \{x_1^{(i)}, x_2^{(i)}, \ldots, x_{\lfloor 0.8 T_i \rfloor}^{(i)}\} \\
\mathbf{x}_{\text{test}}^{(i)} &= \{x_{\lfloor 0.8 T_i \rfloor + 1}^{(i)}, \ldots, x_{T_i}^{(i)}\}
\end{align}

The forecasting methodology varies by model architecture. Traditional models (ARIMA, Gradient Boosting) employ iterative, single-step forecasting to cover the entire test set. For models that benefit from fixed context and horizon windows (Chronos, LLaMA, LLaMA-FT), we define domain-specific windowed forecasting tasks:

\noindent\textbf{1. Finance:} Given context length $C = 20$ and horizon $H = 5$, predict 5 business days using the preceding 4 weeks of adjusted closing prices.

\noindent\textbf{2. Power:} Given context length $C = 1440$ and horizon $H = 360$, predict 6 hours of minutely power levels using the previous day's data.

\noindent\textbf{3. Pedestrian:} Given context length $C = 72$ and horizon $H = 18$, predict 18 hours of pedestrian counts using the previous 3 days of hourly data.

\noindent\textbf{4. Car:} Given context length $C = 8$ and horizon $H = 2$, predict 2 months of car part sales using the previous 8 months of data.

For windowed models, let $\mathbf{x}_{t-C+1:t}^{(i)} = \{x_{t-C+1}^{(i)}, \ldots, x_t^{(i)}\}$ denote the context window and $\hat{\mathbf{x}}_{t+1:t+H}^{(i)} = \{\hat{x}_{t+1}^{(i)}, \ldots, \hat{x}_{t+H}^{(i)}\}$ the predicted horizon. The forecasting process slides this window across the test set to generate predictions for the entire test period.

% \subsection{Notation}

%Throughout this paper, we use the following notation:
%\begin{itemize}
%\item $\mathbf{x}^{(i)}$: Time series from dataset $i$
%\item $T_i$: Length of time series $i$
%\item $\mathbf{x}_{\text{train}}^{(i)}, \mathbf{x}_{\text{test}}^{(i)}$: Training and test partitions
%\item $C, H$: Context length and forecasting horizon (model-dependent)
%\item $\hat{x}_t$: Predicted value at time $t$
%\item $\mathcal{M}$ = MODELS HERE
%\end{itemize}
\section{Methods}
We describe the implementation and evaluation procedure used to generate forecasts, interpret model behavior, and compute rating-based explanations. All forecasting experiments follow the unified data split and rolling forecast protocol defined in Section~\ref{sec:problem-setting}.

\subsection{Data Description}
\label{sec:data}
% For each dataset $D_i \in \mathcal{D}$, we observe a univariate time series $\mathbf{x}^{(i)} = {x_1^{(i)}, x_2^{(i)}, \ldots, x_{T_i}^{(i)}}$ where $T_i$ represents the total length of series $i$. While three datasets consist of continuous-time observations, the financial dataset comprises business daily observations (excluding weekends and holidays), which we treat as continuous for forecasting purposes.
We use the following datasets for training and evaluating the TSFM, where $T_i$ denotes the length of the series as defined in Section \ref{sec:problem-setting}.

\noindent\textbf{1. Finance ($D_{1}$)}: Six parallel financial time-series ($T_i = 250$ for all series) representing daily closing prices from  March 28, 2023, to March 22, 2024 (ensuring no overlap with Chronos's training data). Series are evenly split between technology, finance, and pharmaceutical sectors.

\noindent\textbf{2. Power ($D_{2}$)}: A single univariate series ($T_i = 20,915$) of minute-level energy consumption measurements from Tantiv4 \cite{muppasani2023dataset}.

\noindent\textbf{3. Pedestrian ($D_{3}$)}: 66 hourly pedestrian count series with varying lengths ($576 \leq T_i \leq 96,424$) from Melbourne sensors \cite{peds}.

\noindent\textbf{4. Car Parts ($D_{4}$)}: 2,674 monthly sales series ($T_i = 51$ for all series) with extreme sparsity (75.9\% zero values) \cite{hyndman2008forecasting}.

We preprocess each dataset into a consistent long-format schema with columns \texttt{series\_id}, \texttt{timestamp}, and \texttt{value}. We make no assumptions regarding stationarity, ergodicity, or other statistical properties of the time-series. All datasets are characterized by inherent forecasting difficulty, representing challenging real-world scenarios where traditional statistical assumptions may not hold. These datasets collectively capture a range of forecasting scenarios. 

%\begin{itemize}
%    \item \textbf{Short, clean economic series} (Finance)
%    \item \textbf{Long-range volatile univariate modeling} (Power)
%    \item \textbf{Variable-length multiseries data} (Pedestrian)
%    \item \textbf{Sparse high-dimensional panel data} (Car Parts)
%\end{itemize}

 %—from learning across many short series to capturing dynamics in long or sparse sequences—while also reflecting real-world deployment conditions.

\subsection{Forecasting Procedure}
To generate predictions beyond each model's native horizon, we adopt an \textbf{iterative forecasting} strategy. Models first predict for a window of length equal to their native horizon; these predictions are appended to the historical context and used to forecast the next window. This process repeats until the test period is fully covered. The full prediction history is retained for each series, and only the required number of steps is generated at each iteration to avoid compounding unnecessary error. Two inference setups are used:

\noindent \textbf{Autoregressive:} Iterative rollout until the forecast window is covered.

\noindent  \textbf{Direct:} A single context window is used to predict the entire horizon in one shot.

For Llama and its fine-tuned version, we use the direct setup on the Power and Pedestrian datasets due to their large size, which makes full autoregressive rollouts impractical within our time budget. All other datasets use the autoregressive setup. Domain-specific context ($C$) and horizon ($H$) lengths are detailed in Section \ref{sec:problem-setting}. During training, sliding windows are extracted from the training series. During inference, these windows slide across the test period to cover the full horizon.

\subsection{Models}
\label{sec:models}
We evaluate four models spanning statistical baselines, feature-based ML, and foundation models (Table~\ref{tab:models_used}).

\noindent\textbf{1. ARIMA:} Implemented with \texttt{statsmodels} per series. Data are scaled with a \texttt{RobustScaler} and inverse-transformed after forecasting. Model orders $(p,d,q)$ are selected via grid search constrained by domain-specific maxima; $d$ is chosen using the Augmented Dickey–Fuller and KPSS tests. Seasonal orders $(P,D,Q,s)$ are included when $s>1$ and data length permits, with $s$ inferred from autocorrelation peaks or domain defaults (e.g., $s=5$ for finance, $s=1440$ for power). Grid search considers $p,q \leq 5$, $d \in \{0,1,2\}$. Forecasting is iterative single-step with a rolling window ($w=7$ for finance, $w=1440$ for power). Confidence intervals (95\%) are used for uncertainty quantification.

\noindent\textbf{2. Gradient Boosting:} Implemented with LightGBM (\texttt{lgb.LGBMRegressor}) using engineered features: time-based (hour, day, month), Fourier terms for seasonality, domain-specific statistics (e.g., log returns, volatility, moving averages), and tailored lags. Models are trained independently on the training set only (500 estimators, learning rate 0.05, L1 loss). Forecasting is iterative, with features recalculated after each prediction.

\noindent\textbf{3. Chronos:} A pretrained transformer for time-series, implemented via AutoGluon TimeSeries in ``bolt\_small'' configuration. Data are resampled to known frequencies per domain (hourly, monthly, minutely, business days), using forward-fill interpolation except for car sales, which uses zero-fill to preserve sparsity. All numeric values are cast to float32.

\noindent \textbf{4. Llama/Llama-FT}: \texttt{Meta-Llama-3.1-8B-Instruct} fine-tuned with LoRA adapters on MinMax-scaled windows, where each prompt contains $C$ past values and $H$ future targets. LoRA configuration: $r=16$, $\alpha=32$, dropout $=0.05$; quantization: NF4, bfloat16; optimizer: paged AdamW with cosine LR decay, gradient accumulation, and fp16. We refer to the fine-tuned variant as Llama-FT and the original base model as Llama.

\subsection{Evaluation Metrics}
Model accuracy is evaluated with:
\[
\text{sMAPE} = \frac{100}{n} \sum_{t=1}^{n} \frac{|x_t - \hat{x}_t|}{(|x_t| + |\hat{x}_t|)/2}
\]
\[
\text{MASE} = \frac{\frac{1}{n}\sum_{t=1}^{n}|x_t - \hat{x}_t|}{\frac{1}{m-s}\sum_{t=s+1}^{m}|x_t - x_{t-s}|}
\]
where $n$ is the test length, $m$ the training length, and $s$ the seasonal period ($s=1440$ power, $s=24$ pedestrian, $s=5$ finance, $s=1$ car). Metrics are computed per series and aggregated by domain.

\subsection{Explainability Analysis}
We use SHAP to compute feature attributions. For GBoost, TreeSHAP is applied directly. For Chronos, surrogate LightGBM models approximate outputs for SHAP analysis \cite{raykar2023tsshap}, with fidelity checked via RMSE. TS-MULE (a LIME variant for time-series) is also used, perturbing uniform segments with mean or zero replacement to explain the final forecast value.

\subsection{Rating-Driven Explanations (RDE)}
\label{sec:rde}
\begin{figure}[bh]
    \centering
    \includegraphics[width=0.22\textwidth]{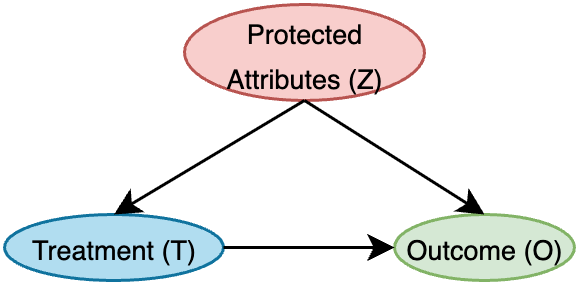}
    \caption{Proposed Generalized Causal Graph.}
    \label{fig:gen-cm}
\end{figure}
The RDE framework~\cite{lakkaraju2025h-xai} integrates causal reasoning with metrics to explain model behavior under different conditions. The generalized causal graph used for our experiments is shown in Figure \ref{fig:gen-cm}. A causal model specifies relationships between treatment $T$, outcome $O$, and protected attribute $Z$. If $Z$ confounds $T$ and $O$, deconfounding is required. We focus on:
\begin{itemize}
    \item \textbf{Weighted Rejection Score (WRS):} Measures statistical differences in $O$ across $Z$ groups via multiple $t$-tests, weighted by significance level. Higher WRS = more $Z$ sensitivity.
    \item \textbf{Average Treatment Effect (ATE):} Estimates the causal effect of $T$ on $O$ using G-computation \cite{robins1986new} to account for confounding.
\end{itemize}
More details on the metrics are provided in Section \ref{app:rde} in the appendix. The workflow (Figure~\ref{fig:rde-workflow}) takes a user query or a hypothesis (as shown in Tables \ref{tab:ate} and \ref{tab:wrs}), identifies $T$, $O$, and $Z$, selects a metric, computes metrics, and compares them to random and biased baselines to contextualize findings.

\section{Experiments and Results}
\begin{figure}[htbp]
    \centering
    \begin{subfigure}[b]{0.42\textwidth}
        \centering
        \includegraphics[width=\linewidth]{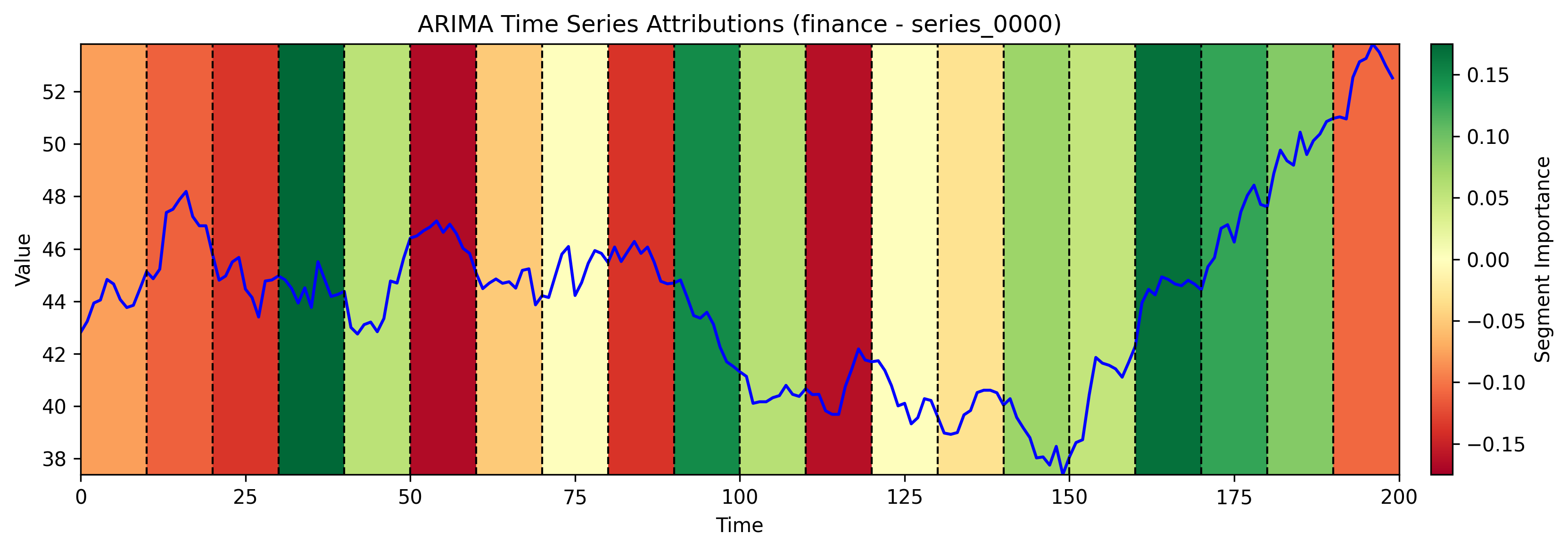}
        \caption{Finance series\_0}
        \label{fig:ARIMA_LIME_finance0}
    \end{subfigure}
    \hfill
    \begin{subfigure}[b]{0.42\textwidth}
        \centering
        \includegraphics[width=\linewidth]{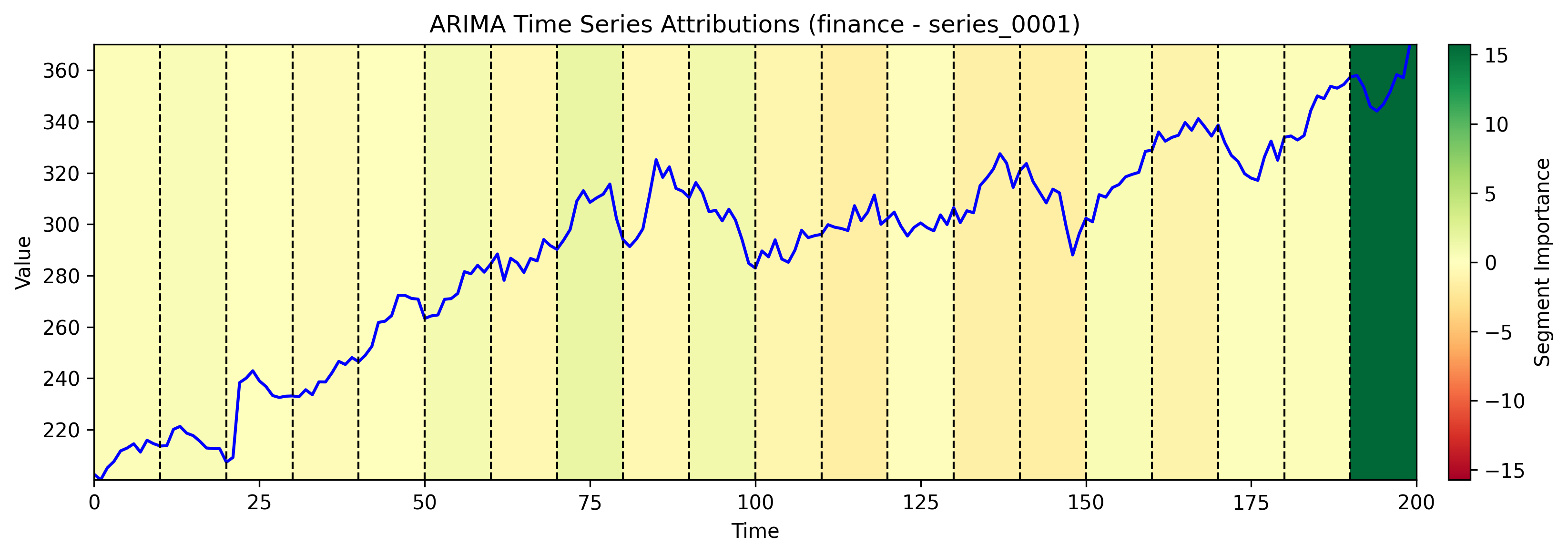}
        \caption{Finance series\_1}
        \label{fig:ARIMA_LIME_finance1}
    \end{subfigure}
    \hfill
    \begin{subfigure}[b]{0.42\textwidth}
        \centering
        \includegraphics[width=\linewidth]{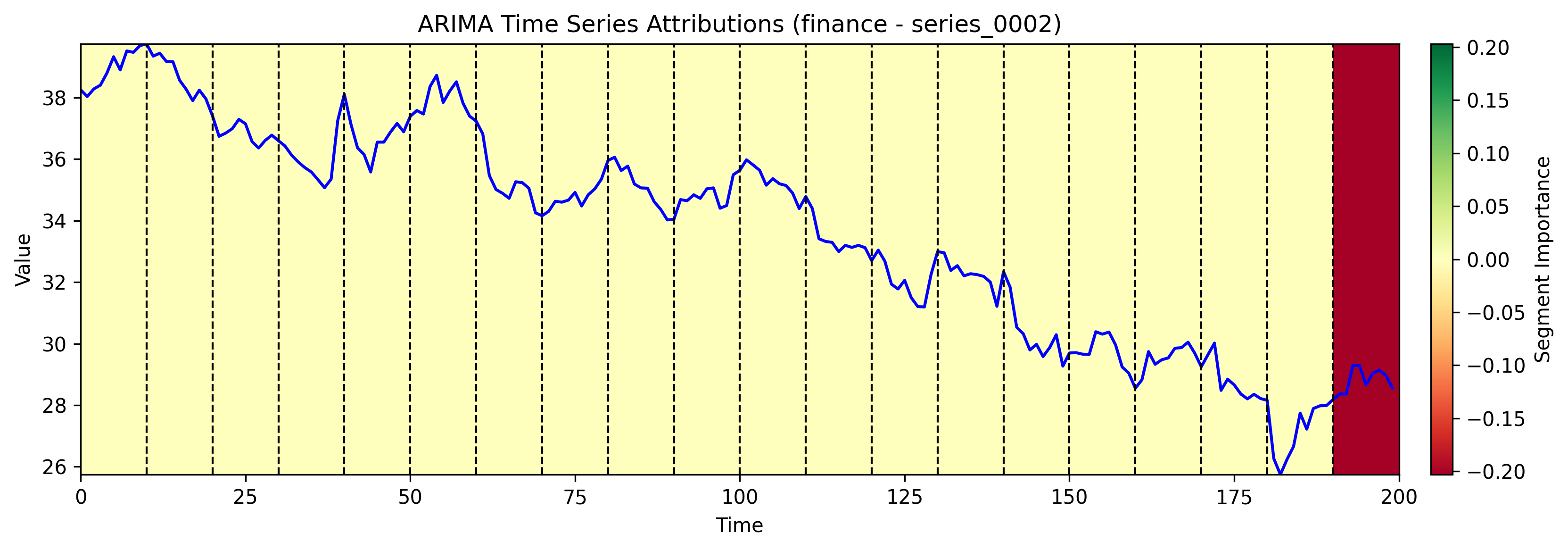}
        \caption{Finance series\_2}
        \label{fig:ARIMA_LIME_finance2}
    \end{subfigure}
    \hfill
    \begin{subfigure}[b]{0.42\textwidth}
        \centering
        \includegraphics[width=\linewidth]{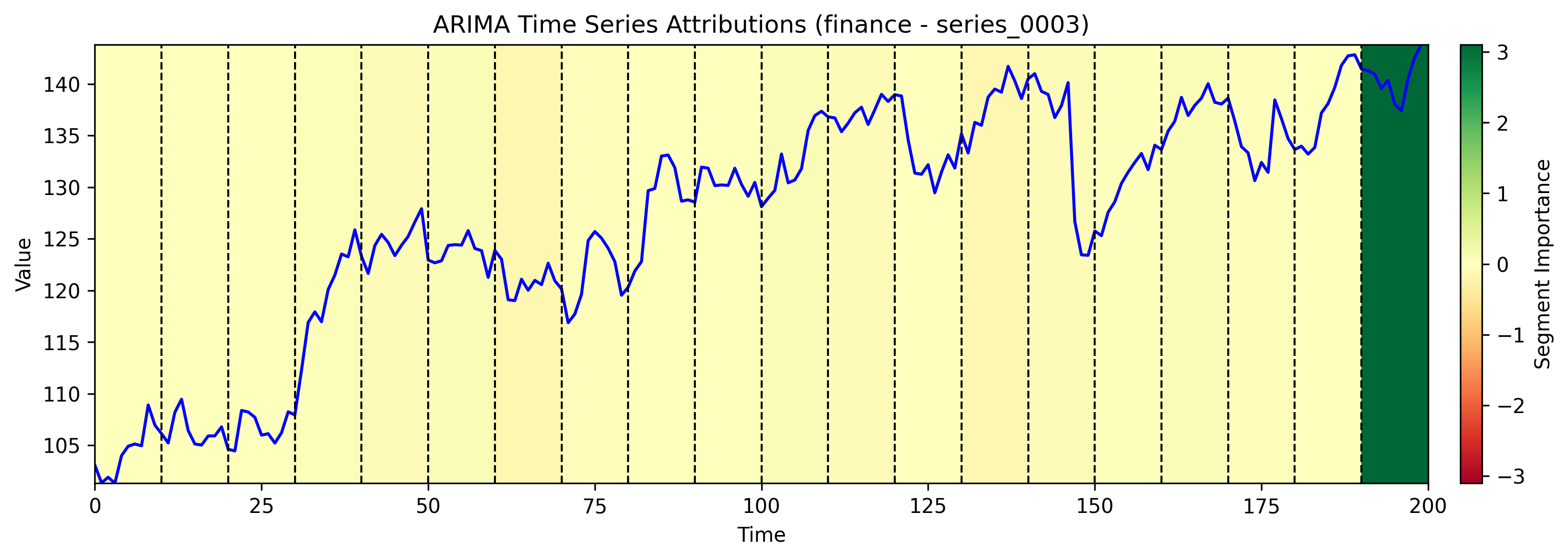}
        \caption{Finance series\_3}
        \label{fig:ARIMA_LIME_finance3}
    \end{subfigure}
    \hfill
    \begin{subfigure}[b]{0.42\textwidth}
        \centering
        \includegraphics[width=\linewidth]{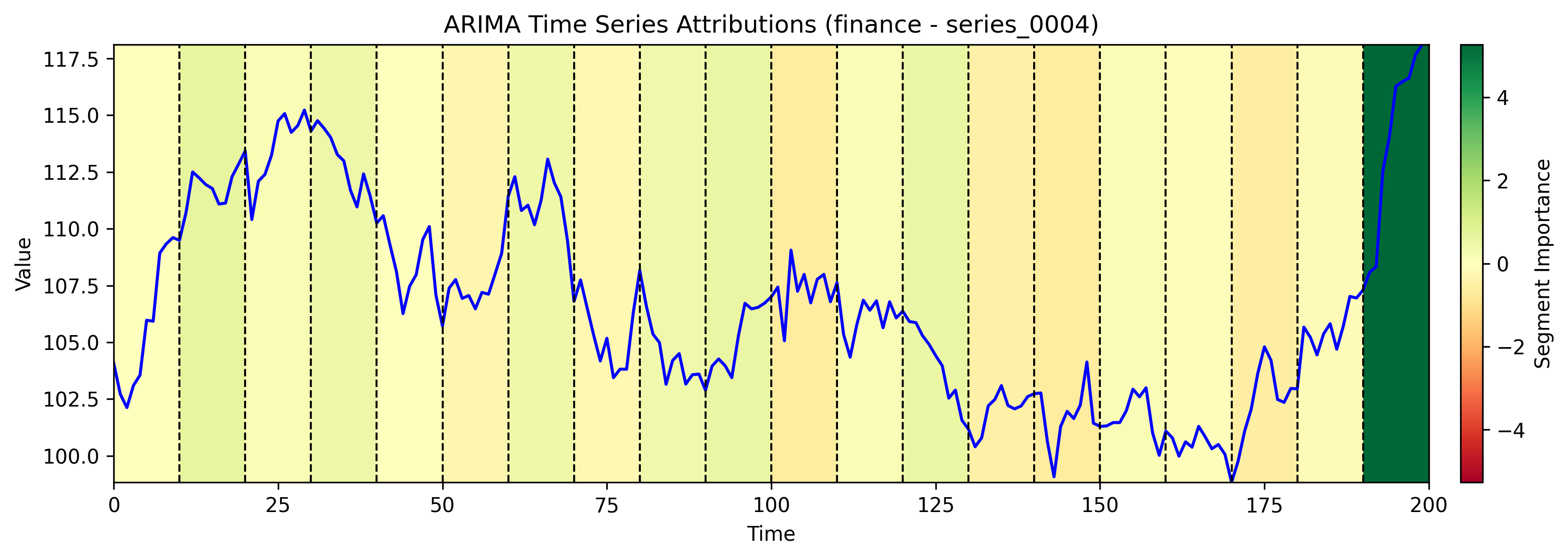}
        \caption{Finance series\_4}
        \label{fig:ARIMA_LIME_finance4}
    \end{subfigure}
    \hfill
    \begin{subfigure}[b]{0.42\textwidth}
        \centering
        \includegraphics[width=\linewidth]{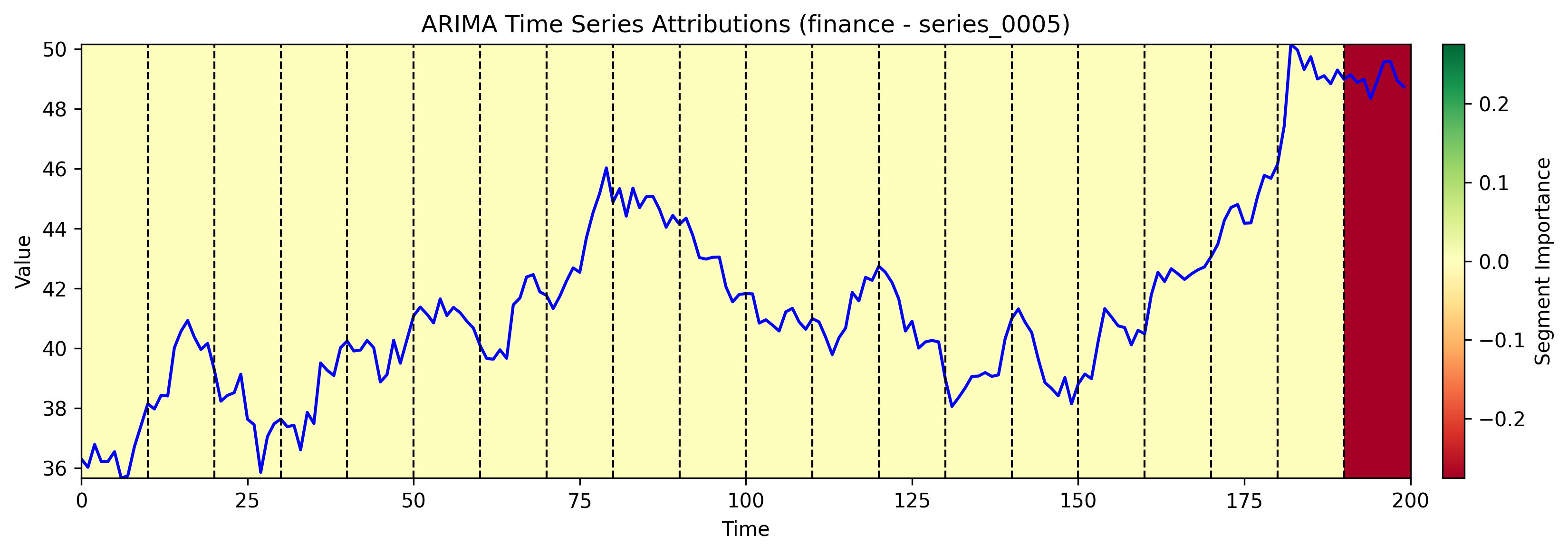}
        \caption{Finance series\_5}
        \label{fig:ARIMA_LIME_finance5}
    \end{subfigure}
    \caption{LIME plots for ARIMA model in finance using 20 uniform segments, local-mean replacement, and 200 samples. Figures show the distribution of feature importance for each time-series. Panels (c) and (f) show very little feature importance ($<$ 0.000001) assigned to values to before the last segment, while panels (a), (b), (d), and (e) show varying feature importance in all segments.}
    \label{fig:ARIMA_lime_values}
\end{figure}

In this section, we detail the experimental setup used to address the research questions stated in Section \ref{sec:introduction}.
\subsection{RQ1: When forecasting models succeed or fail?} 
\begin{table*}
    \centering
    \small  % Slightly smaller font for better fit
    \begin{tabular}{l l r@{\,±\,}l r@{\,±\,}l r@{\,±\,}l r@{\,±\,}l r@{\,±\,}l r@{\,±\,}l}
        \toprule 
        \textbf{Domain} & \textbf{Metric} & \multicolumn{2}{c}{\textbf{ARIMA}} & \multicolumn{2}{c}{\textbf{GBoost}} & \multicolumn{2}{c}{\textbf{Chronos*}} & \multicolumn{2}{c}{\textbf{LLaMa}} & \multicolumn{2}{c}{\textbf{LLaMa-FT}}\\
        \midrule
        \textbf{Finance} & MASE 
                         & 7.69 & 6.56 
                         & \textbf{4.32} & \textbf{3.00} 
                         & 5.48 & 5.44 
                         & 88.06 & 23.88  
                         & 8.92 & 6.13  \\
                         & sMAPE 
                         & 5.90 & 6.37 
                         & 10.27 & 7.37 
                         & \textbf{5.30} & \textbf{5.15} 
                         & 169.03 & 47.91 
                         & 8.64 & 5.99 \\
        \midrule
        \textbf{Power} & MASE 
                       & 2.89 & 0.00 
                       & \textbf{0.84} & \textbf{0.00} 
                       & 2.33 & 0.00 
                       & 40.06 & 0.00
                       & 14.07 & 0.00 \\
                       & sMAPE 
                       & 22.08 & 0.00 
                       & \textbf{6.86} & \textbf{0.00} 
                       & 8.84 & 0.00 
                       & 17.75  & 0.00
                       & 6.04  & 0.00  \\
        \midrule
        \textbf{Pedestrian} & MASE 
                            & 1.86 & 7.77 
                            & \textbf{1.00} & \textbf{0.39} 
                            & 1.47 & 0.67* 
                            & 3.21 & 1.57 
                            & 1.77 & 0.77\\
                           & sMAPE 
                           & 214.41 & 27.2 
                           & \textbf{50.64} & \textbf{11.78}
                           & 90.98 & 44.77* 
                           & 107.94  & 61.36 
                           & 72.85  & 13.46 \\
        \midrule    
        \textbf{Car Parts} & MASE 
                           & 1.38 & 3.40 
                           & \textbf{0.77} & \textbf{1.32} 
                           & \multicolumn{2}{c}{$\infty$}    
                           & \multicolumn{2}{c}{$\infty$}
                           & \multicolumn{2}{c}{$\infty$}  \\
                          & sMAPE 
                          & \multicolumn{2}{c}{$\infty$} & \textbf{49.97} & \textbf{52.80}
                          & 175.03 & 37.56* 
                          & 142.76 & 89.46
                          & 138.53 & 89.04 \\
        \bottomrule
    \end{tabular}
    \caption{Forecasting performance comparison. MASE and sMAPE are reported as (mean $\pm$ std. dev.). sMAPE is reported as percentage (\%). GBoost $\rightarrow$ Gradient Boosting. Best results are in \textbf{bold} (lower value is better). {\bf Across domains, GBoost exhibits best MASE and sMAPE with the exception of sMAPE finance}. 
    *During the initial training and evaluation of Chronos \cite{ansari2024chronos}, the model was exposed to the pedestrian and car parts dataset in an experimental setup different than our own.}
    \label{tab:model_performance}
\end{table*}
\noindent \textbf{Experimental Setup:} To answer this question, we collected complementary performance metrics sMAPE and MASE through forecasting experiments as detailed in Sections 3 and 4. 

\noindent \textbf{Results:} Table \ref{tab:model_performance} summarizes forecasting performance across four domains using MASE and sMAPE, with lower values indicating better accuracy. Gradient Boosting (GBoost) generally achieved the strongest overall performance, obtaining the lowest MASE in finance (4.32 ± 3.00), power (0.84 ± 0.00), pedestrian (1.00 ± 0.39), and car parts (0.77 ± 1.32). Chronos performed competitively in certain cases, delivering the best sMAPE for finance (5.30 ± 5.15 \%) and strong results for power (8.84 \%), though its pedestrian accuracy lagged behind GBoost. ARIMA yielded reasonable performance in some domains but was consistently outperformed by GBoost, particularly in power and pedestrian datasets where the latter reduced MASE by more than half.

The LlaMA model, without fine-tuning, struggled across domains, producing large errors, especially in finance (MASE = 88.66) and pedestrian datasets, while fine-tuning (LlaMA-FT) reduced errors substantially but did not surpass GBoost or Chronos in any metric. In the car parts domain, only GBoost and ARIMA results are available for MASE due to Chronos limitations, while sMAPE is missing for ARIMA and GBoost. Across all domains, the standard deviation of results highlights the variability in model stability, with GBoost consistently showing both strong mean performance and relatively low variance, reinforcing its robustness across heterogeneous time-series forecasting tasks.

\noindent\textbf{Interpretation:} Forecasting models succeed or fail based on how well their assumptions and architectures align with the data's statistical and structural properties. Gradient Boosting excels across most domains, leveraging engineered time-based and domain-specific features to overcome challenges like inconsistent seasonality, sparsity, or long sequences. In contrast, ARIMA performs well on short, stationary series with strong autocorrelations but struggles with high-frequency or nonstationary data (e.g., minute-level power consumption) due to its linearity. Chronos generalizes well, achieving strong sMAPE in finance and competitive results in power. However, like other models, its advantage diminishes with irregular seasonality or extreme sparsity (e.g., car parts, pedestrian datasets), where feature-driven methods like Gradient Boosting outperform it. General-purpose FM, Llama, is highly sensitive to sequence length, sparsity, and distributional shifts; without fine-tuning, it performs poorly. While Llama's low sMAPE suggests good proportional accuracy, its high MASE reveals larger absolute errors than the naive baseline, which performed well due to minimal inter-step variation (Table~\ref{tab:model_performance}).

\noindent\textbf{Conclusion:} Ultimately, model success hinges on the alignment between inductive bias and dataset characteristics (frequency, length, regularity, sparsity, seasonality). Domain knowledge integration remains the most critical factor in time-series forecasting.
\subsection{RQ2: What do common XAI methods tell us about why forecasting models succeed or fail?} %Though popular XAI methods like LIME and SHAP can help users understand the effects and benefits of quality feature engineering, they do more to justify model predictions than to inform user decision making. 
\noindent \textbf{Experimental Setup:} To answer this question, we applied variations of SHAP and LIME as detailed in sections 3 and 4. 

\begin{figure}[htbp]
    \centering
    \includegraphics[width=0.5\linewidth]{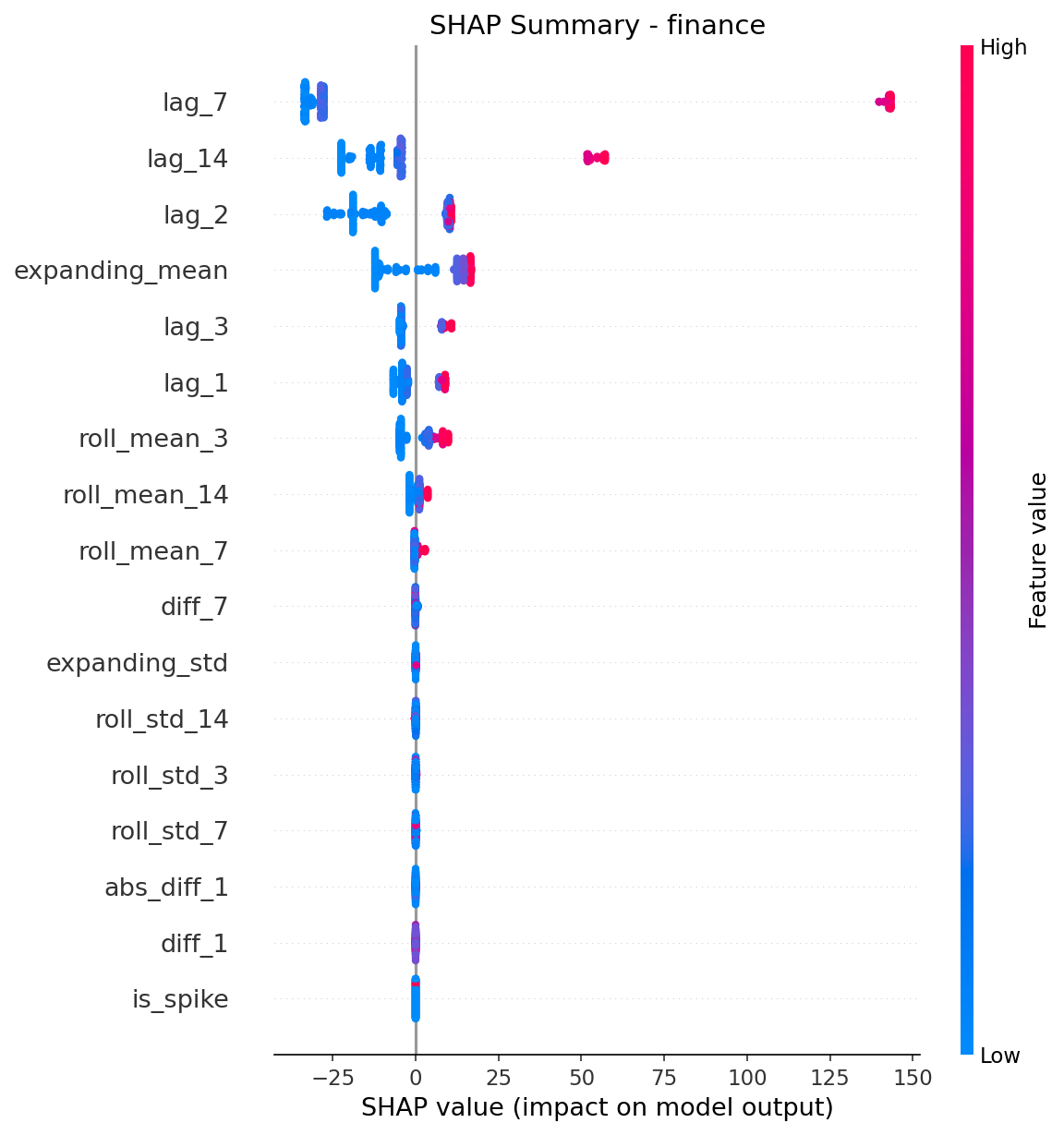}
    \caption{SHAP values for the Chronos Surrogate model (finance domain). The full set of domain-specific results is provided in Figure~\ref{fig:chronos_shap_values} (Appendix).}
    \label{fig:chronos_shap_finance}
\end{figure}

\noindent \textbf{TsSHAP - Chronos Surrogate Model:}
In the \textbf{Finance} domain, lag features comprise 4 of the top 5 features, with expanding mean being the only non-lag feature. This, combined with the Surrogate's performance metrics found in the appendix Table \ref{tab:surrogate_performance} shows that Chronos and its surrogate captures trends despite no domain-specific features. The plot presents a clear directional pattern: high historical values boost predictions, low values suppress them. Surrogate maintains GBoost's lag sensitivity but replaces engineered features with statistical aggregates. 

\noindent\textbf{Interpretation:} When placed in conversation with the other domains found in Table \ref{fig:chronos_shap_values} (in appendix), Chronos's surrogate reveals a fundamental tension, while capable of detecting simple trends (finance lags), they default to statistical baselines (means/standard deviations) when faced with complex patterns. This suggests pretrained models may \textit{recognize} temporal structures but struggle to \textit{leverage} them without domain-specific feature engineering. The surrogate's performance floor becomes evident in challenging domains (car/power), where it reverts to predicting averages rather than meaningful values. 

\begin{figure}[htbp]
    \centering
    \includegraphics[width=0.75\linewidth]{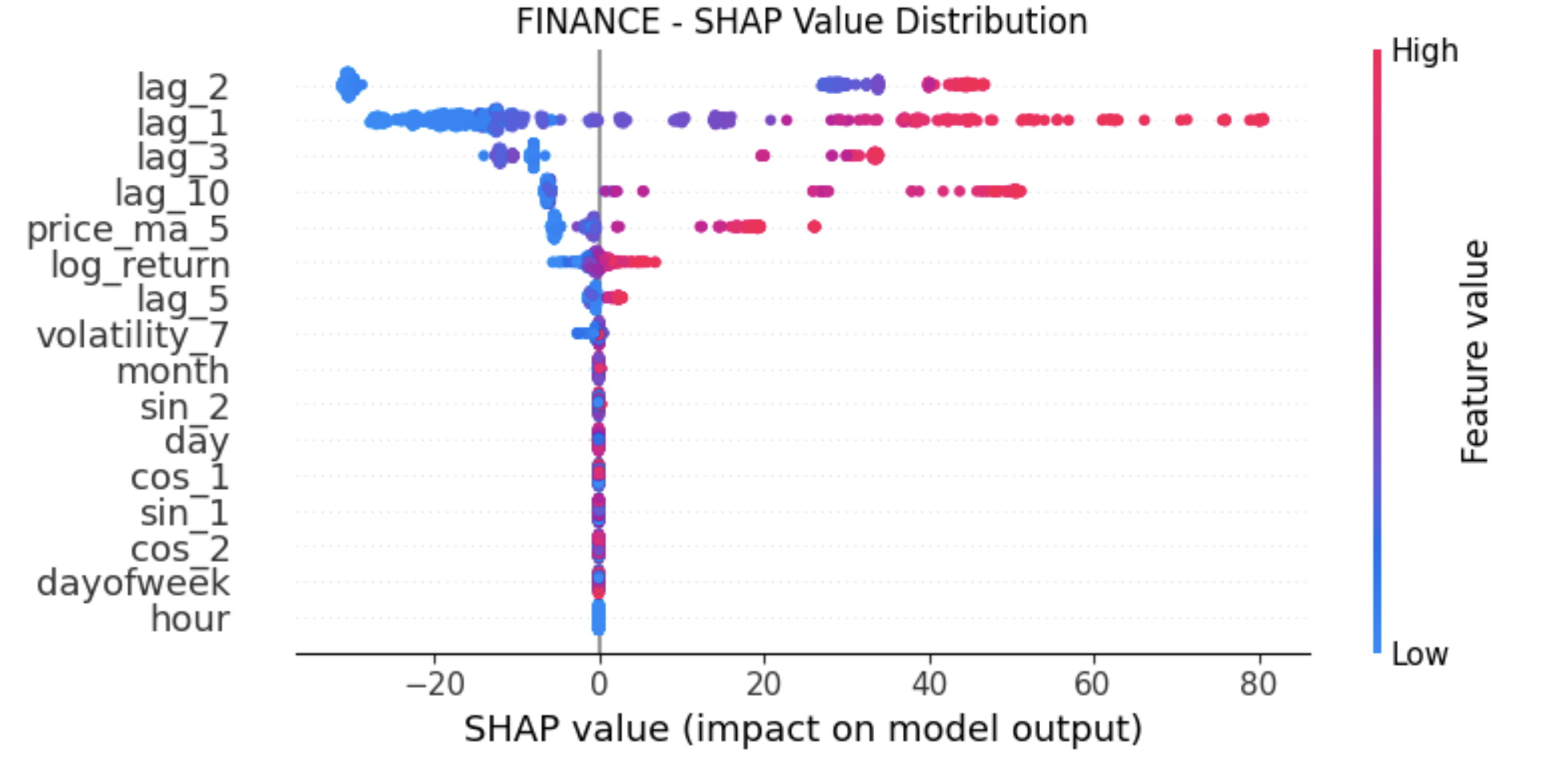}
    \caption{SHAP values for the Gradient Boosting model (finance domain). The full set of domain-specific results is provided in Figure~\ref{fig:gb_shap_values} (Appendix).}
    \label{fig:gb_shap_finance}
\end{figure}

\noindent\textbf{TreeSHAP - Gradient Boosting:}
In \textbf{Finance}, lagged values constitute 4 of the top 5 features, alongside a 5-day moving average. This confirms the trend-driven nature of financial markets, where price movements exhibit momentum effects. Calendar features (day/month) showed negligible importance, suggesting market reactions depend more on events than temporal context. The dominance of engineered features over raw lags highlights how domain-specific transformations can capture meaningful patterns. 

\noindent\textbf{Interpretation:} In the other 3 domains (Appendix) we see a similar pattern, while lagged values provide a universal foundation, each domain's unique physics (market trends, human schedules, periodicity, or sparsity patterns) dictate which features ultimately drive predictive power.

\begin{figure}[htbp]
    \centering
    \begin{subfigure}[b]{0.23\textwidth}
        \centering
        \includegraphics[width=\linewidth]{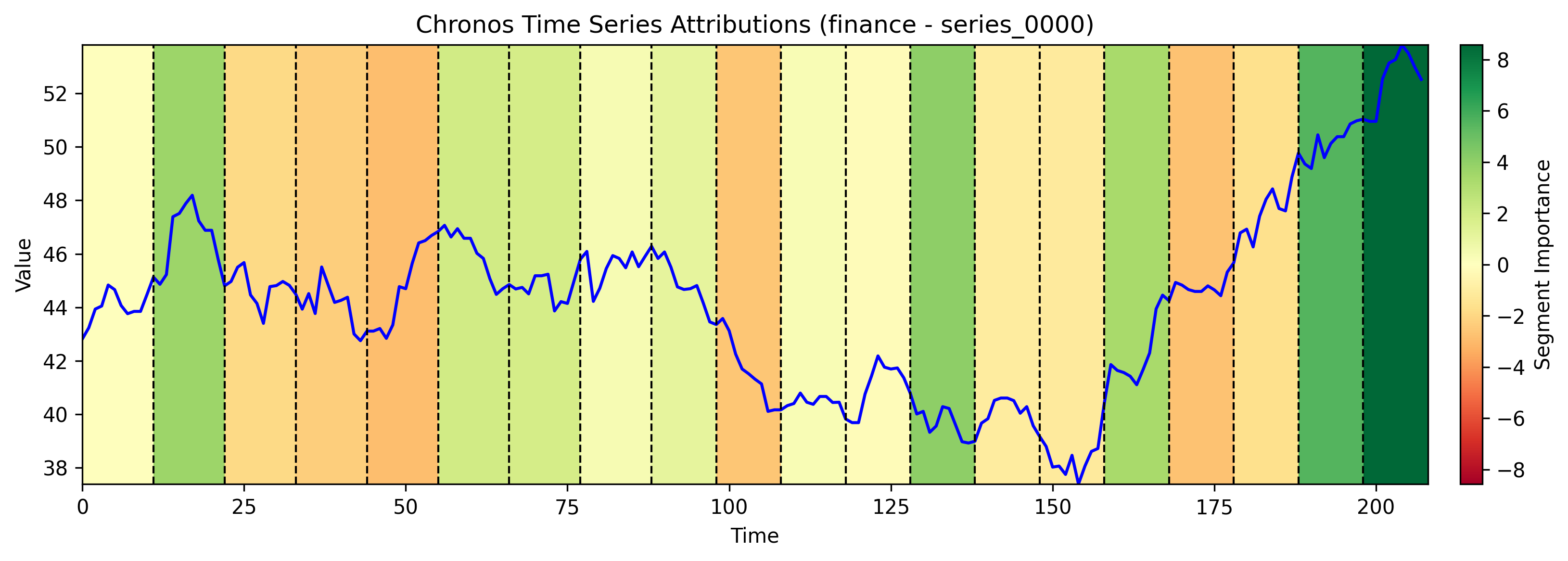}
        \caption{Finance series\_0}
        \label{fig:Chronos_LIME_finance0}
    \end{subfigure}
    \hfill
    \begin{subfigure}[b]{0.23\textwidth}
        \centering
        \includegraphics[width=\linewidth]{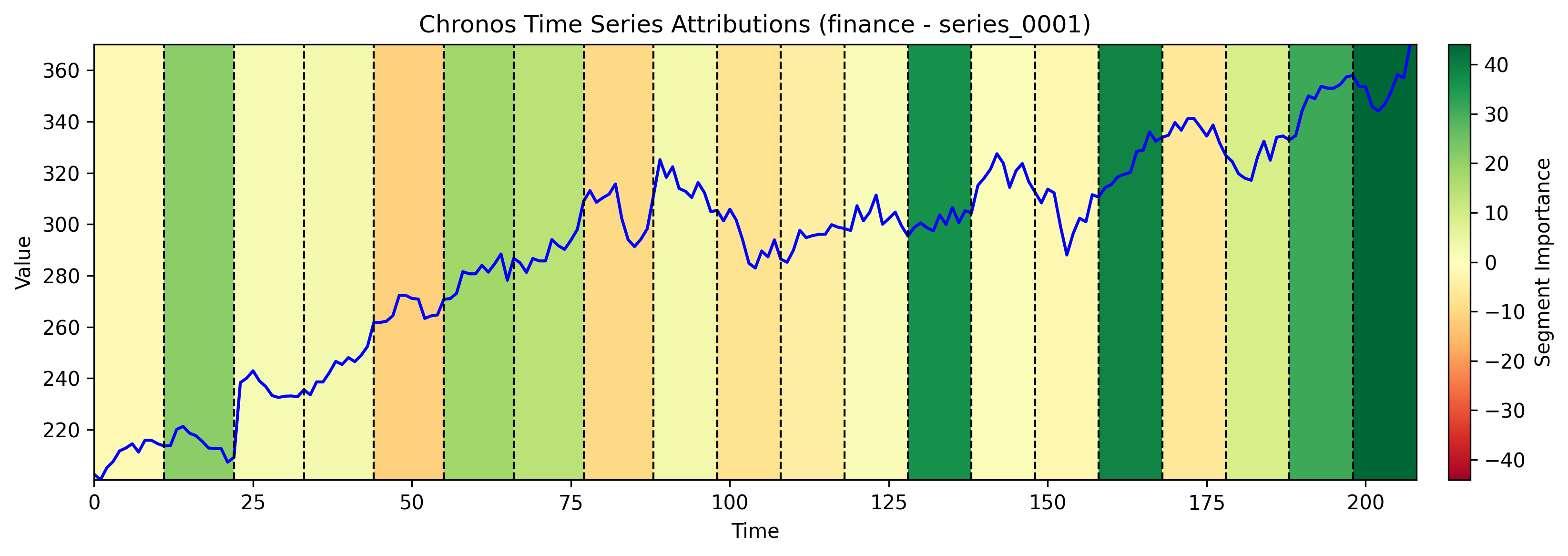}
        \caption{Finance series\_1}
        \label{fig:Chronos_LIME_finance1}
    \end{subfigure}
    \hfill
    \begin{subfigure}[b]{0.23\textwidth}
        \centering
        \includegraphics[width=\linewidth]{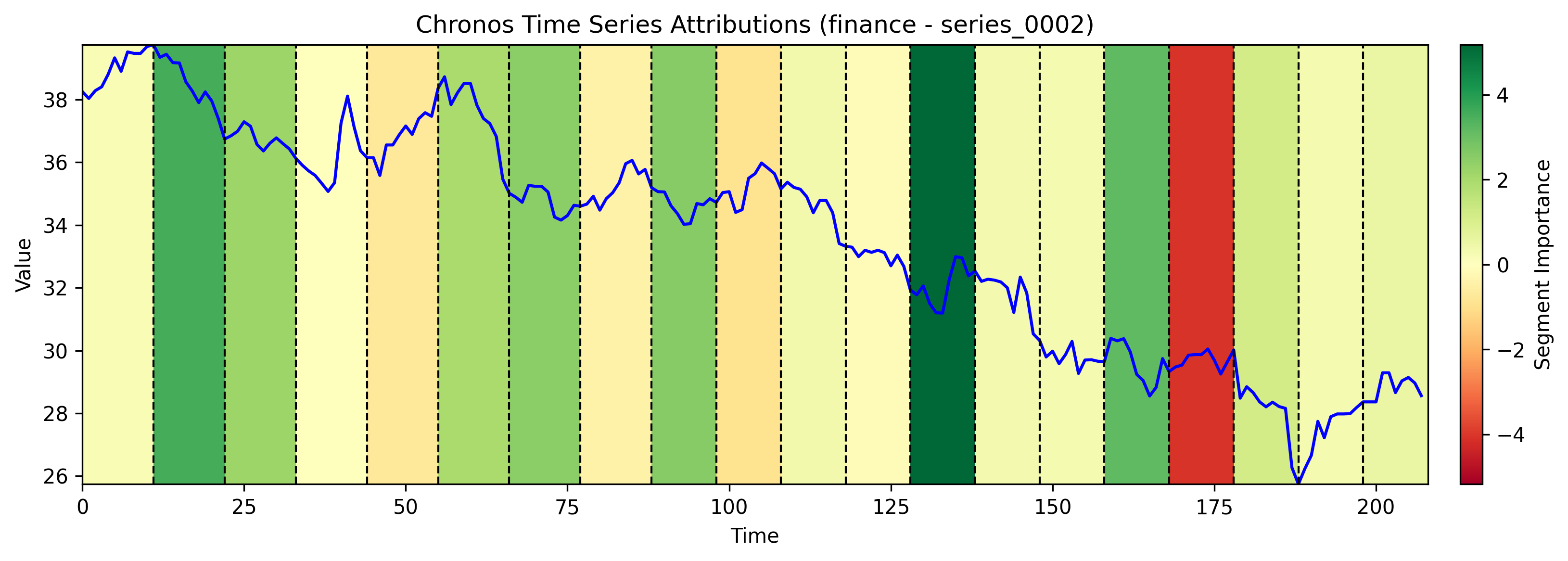}
        \caption{Finance series\_2}
        \label{fig:Chronos_LIME_finance2}
    \end{subfigure}
    \hfill
    \begin{subfigure}[b]{0.23\textwidth}
        \centering
        \includegraphics[width=\linewidth]{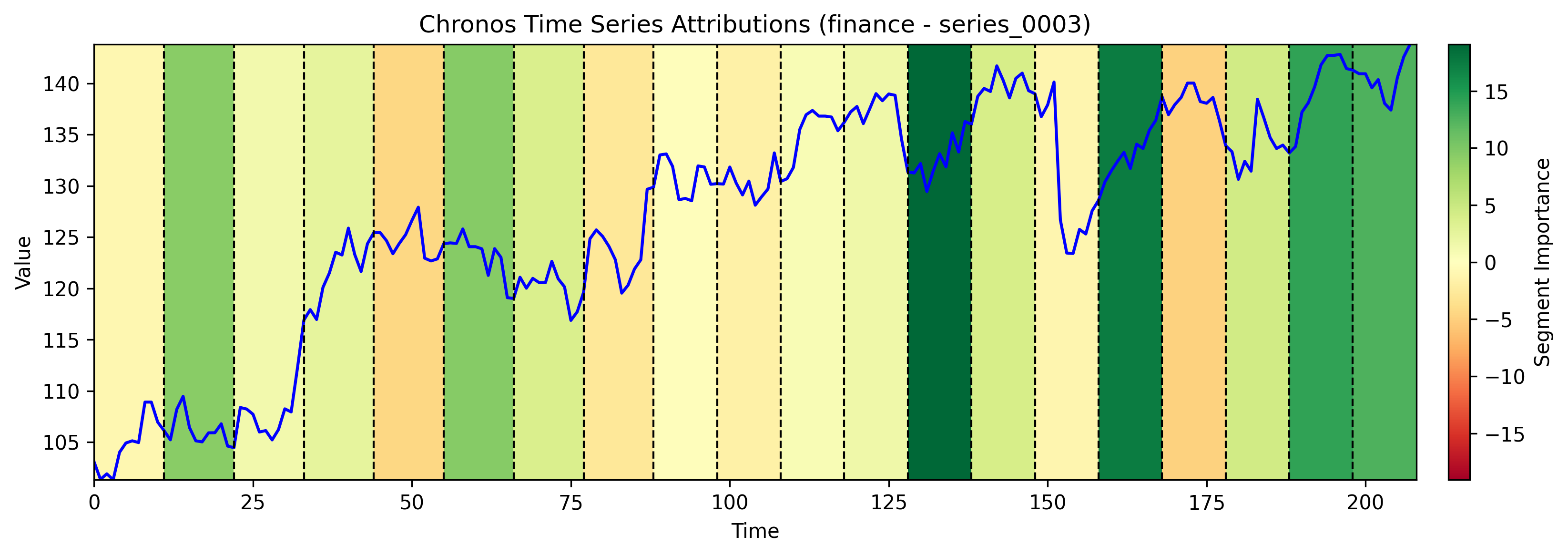}
        \caption{Finance series\_3}
        \label{fig:Chronos_LIME_finance3}
    \end{subfigure}
    \hfill
    \begin{subfigure}[b]{0.23\textwidth}
        \centering
        \includegraphics[width=\linewidth]{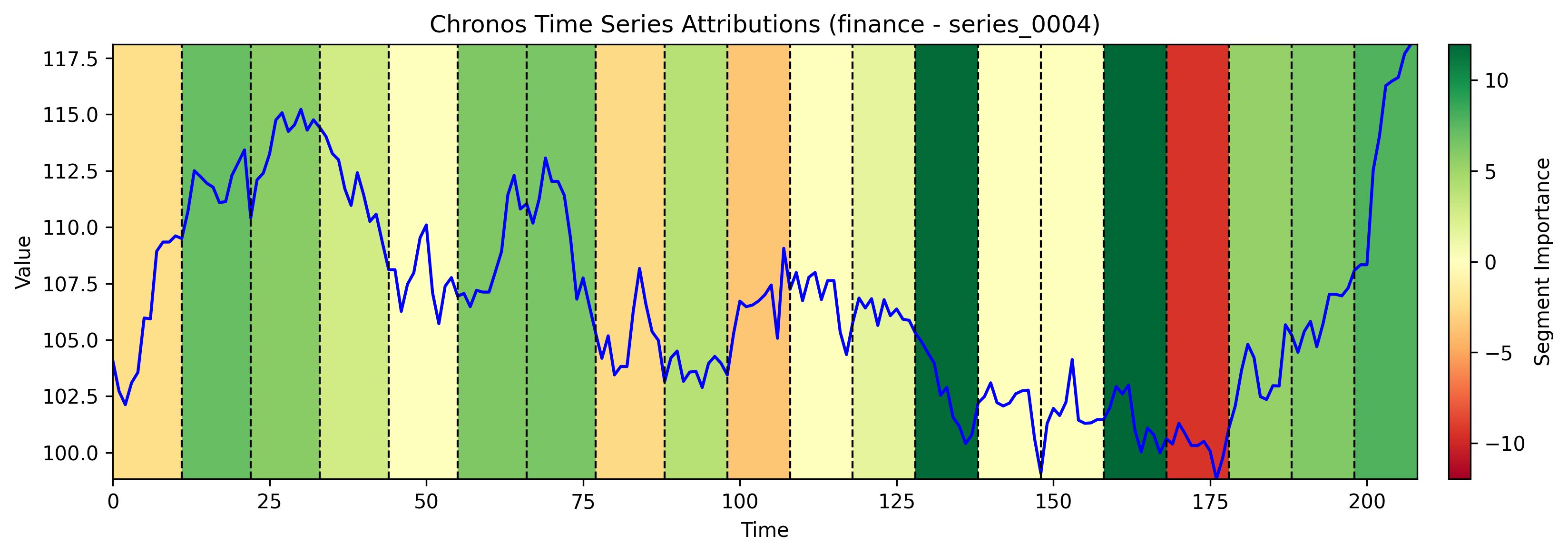}
        \caption{Finance series\_4}
        \label{fig:Chronos_LIME_finance4}
    \end{subfigure}
    \hfill
    \begin{subfigure}[b]{0.23\textwidth}
        \centering
        \includegraphics[width=\linewidth]{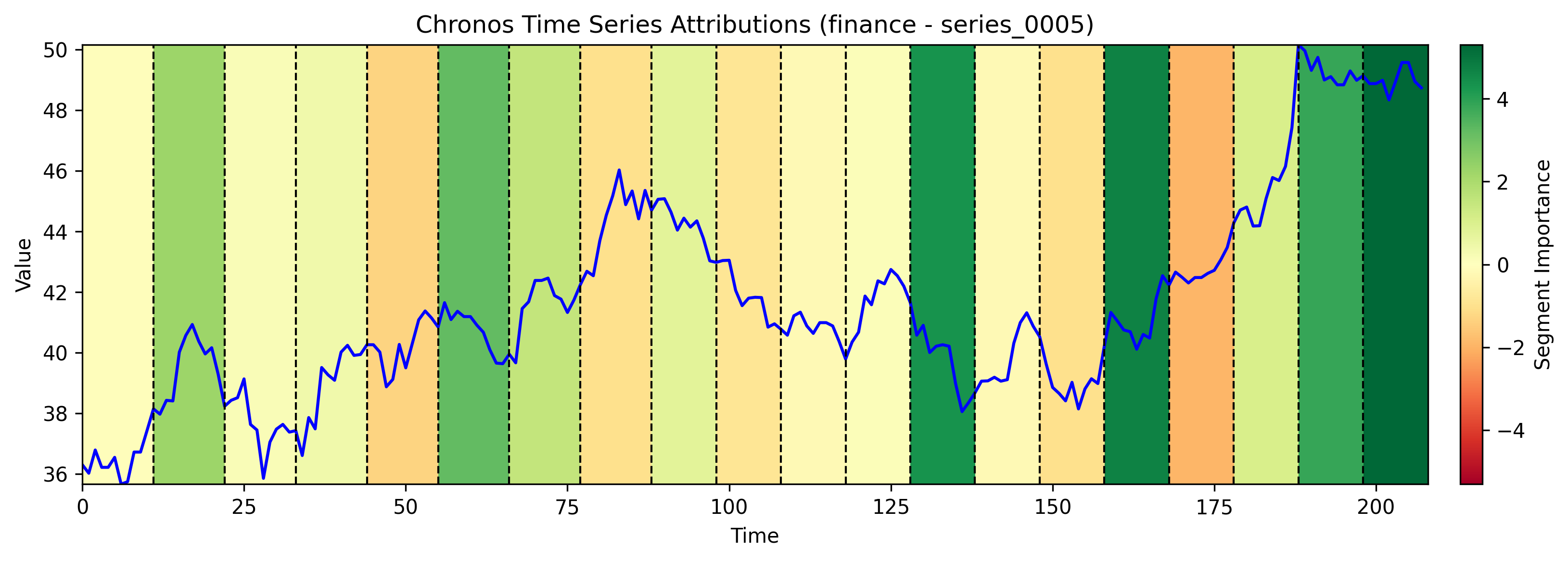}
        \caption{Finance series\_5}
        \label{fig:Chronos_LIME_finance5}
    \end{subfigure}
    \caption{LIME plots for Chronos model in finance using 20 uniform segments, zero replacement, and 200 samples. Figures show the distribution of feature importance for each time-series, revealing a relatively balanced distribution, with slight bias towards the later segments.}
    \label{fig:chronos_lime_finance_values}
\end{figure}
\noindent\textbf{Segment Based LIME - Chronos:}
In \textbf{Finance}, figure \ref{fig:chronos_lime_finance_values} clearly shows that earlier points in the time-series have significant effects on the final value of the forecast, indicating that Chronos is picking up on patterns in the data and using them to determine the direction of the overall forecast. Since zero-replacement is being used, it is expected that all final segments contribute positively towards the final forecast value. A clear pattern emerged with high feature importance being placed on the segments corresponding the 130-140th and 150-160th points in the time-series. The segments which bring down the value of the final forecast do not follow any obvious patterns, indicating that this explanation method may be a better diagnostic tool, rather than a tool for informed decision making.

%In \textbf{Pedestrian}, the explanation shows a similar distribution of feature importance, but has a noticable bias of negative feature importance for earlier segments and higher feature importance for later ones. The explanation also shows that the majority of segments which fully capture high traffic days with peaks at rush hours significantly drive the forecast down.
%\xander{@John move the above commented section to appendix with relevant figure}

\noindent\textbf{Segment Based LIME - ARIMA:}
The LIME explanations for ARIMA in \textbf{finance} clearly demonstrate its strong dependence on the most recent time segments, which aligns  with the model's autoregressive structure. This behavior is inherent to ARIMA's design - the autoregressive component explicitly prioritizes recent observations when generating forecasts, creating the characteristic pattern of exponentially decaying influence as we look further back in time. Figures \ref{fig:ARIMA_LIME_finance2} and \ref{fig:ARIMA_LIME_finance5} show that for the last time point in the forecast, none of the segments other than the last had much of an effect on the output. Visually, figure \ref{fig:ARIMA_LIME_finance0} looks like an outlier, but only shows a lower attribution to the final segment. The values of all other segments are consistent with the distributions seen in figures \ref{fig:ARIMA_LIME_finance1}, \ref{fig:ARIMA_LIME_finance3}, and \ref{fig:ARIMA_LIME_finance4}. 

% \subsection{LLaMA/LLaMA-FT \& Surrogates}
% \xander{Kausik, fill in when ready. Ignore surrogates if timeline dictates}

\subsection{RQ3: How can rating augment our understanding of when and why forecasting models succeed or fail?}

\begin{table}[t]
\tiny
\centering
\setlength{\tabcolsep}{8pt}
\begin{tabular}{@{}llcc@{}}
\toprule
\textbf{Dataset} & \textbf{Model} & \textbf{ATE (lower is better)} & \textbf{Rating} \\
\midrule
\multirow{4}{*}{Cars} 
  & GBoost      & 0.22 & 1 \\
  & ARIMA   & 0.26  & 2 \\
  & Llama-FT & 0.46 & 3 \\
  & Llama      & 0.47  & 4 \\
\midrule
\multirow{4}{*}{Pedestrian} 
  & GBoost      & 320.50 & 1 \\
  & Llama-FT & 500.30 & 2 \\
  & Llama      & 975.48  & 3 \\
  & ARIMA         & 2803.08  & 4 \\
\midrule
\multirow{4}{*}{Finance} 
  & Chronos   & 13.64  &  1\\
  & ARIMA         & 18.19  & 2 \\
  & Llama-FT & 20.20 & 3 \\
  & GBoost      & 28.76 & 4 \\
  & Llama      & 106.02  & 5 \\
\bottomrule
\end{tabular}
\caption{Hypothesis 1: After accounting for the protected attribute, \textbf{the model makes different levels of error depending on which series (car part, company, or sensor) is being predicted}. This shows whether the model is consistent across all series or performs better for some than others.}
\label{tab:ate}
\end{table}

\begin{table}[t]
\tiny
\centering
\setlength{\tabcolsep}{8pt}
\begin{tabular}{@{}llcc@{}}
\toprule
\textbf{Dataset} & \textbf{Model} & \textbf{WRS (lower is better)} & \textbf{Rating} \\
\midrule
\multirow{4}{*}{Cars} 
  & ARIMA    &  0.56 & 1  \\
  & Llama-FT &  0.60 &  2\\
  & Llama      & 0.63 & 3 \\
  & GBoost      &  0.66 &  4 \\
\midrule
\multirow{4}{*}{Pedestrian} 
  & Llama-FT & 0.39 &  1\\
  & Llama      & 0.49 &  2\\
  & ARIMA         & 0.83 & 3 \\
  & GBoost      & 0.87 & 4 \\
\midrule
\multirow{4}{*}{Finance} 
  & Llama      & 0.27  & 1 \\
  & Chronos   & 0.66  &  2\\
  & ARIMA         & 0.75  & 3 \\
  & Llama-FT & 0.85 & 4 \\
 & GBoost      & 0.85 & 4 \\
\bottomrule
\end{tabular}
\caption{Hypothesis 2: Residual distributions differ across protected groups, meaning the \textbf{model makes systematically different errors depending on the month of the year (cars, finance) or the day of the week (pedestrian)}. This matters because such sensitivity can cause the model to be reliable in some periods but noticeably less accurate in others, leading to uneven performance for the user.}
\label{tab:wrs}
\end{table}

\noindent \textbf{Experimental Setup:}
We analyze the same models as in Table~\ref{tab:model_performance} (ARIMA, GBoost, Chronos, LLaMA, LLaMA-FT). Accuracy is reported using \textit{MASE} and \textit{sMAPE}. To examine \textit{how} errors vary, we compute two RDE metrics, previously defined in Section~\ref{sec:rde}, on the Cars, Pedestrian, and Finance datasets. The Power dataset was excluded from the RDE analysis as it contains only a single series with minutely observations, making the definition of a meaningful protected attribute infeasible:

\noindent \textbf{ATE} (Table~\ref{tab:ate}): Estimated effect of the \textit{series identifier} ($T$) on the residual ($O$), after adjusting for the protected attribute ($Z$): month for Cars/Finance, day-of-week for Pedestrian. Higher values indicate greater series-specific error (e.g., part type in Cars, company in Finance, sensor in Pedestrian).

\noindent \textbf{WRS} (Table~\ref{tab:wrs}): Measures differences in residual distributions across protected groups ($Z$): month for Cars/Finance, day-of-week for Pedestrian. Lower values indicate less sensitivity of errors to these groups.

\noindent \textbf{Interpretation:} GBoost's strong accuracy (Table~\ref{tab:model_performance}) coincides with \textit{low ATE} in Cars and Pedestrian (Table~\ref{tab:ate}), errors are relatively uniform across series after adjustment, but \textit{high WRS} across all three datasets (Table~\ref{tab:wrs}). \textbf{Thus, GBoost can be accurate on average yet seasonally/day-wise sensitive, producing uneven errors across months or weekdays.} In Finance, Chronos achieves the lowest ATE (errors consistent across series) but mid-range WRS, while Llama attains the lowest WRS (errors consistent across months/days) but weak ATE and lower accuracy, different models control different sources of variability.
The SHAP summaries (Figures~\ref{fig:chronos_shap_values}, \ref{fig:gb_shap_values}) show heavy reliance on lag/expanding statistics; LIME plots (Figures~\ref{fig:chronos_lime_finance_values}, \ref{fig:ARIMA_lime_values}) often emphasize later segments of the context window. Ratings translate these attribution patterns into measurable consequences: when a model relies on features that co-vary with month/day, \textbf{WRS increases}; when it overfits to series-specific patterns, \textbf{ATE increases}.

\noindent \textbf{Conclusion:} Ratings provide the missing link between forecasting accuracy and feature attributions. ATE captures \textit{series-dependent error} after adjusting for potential confounders, while WRS captures \textit{error differences across months or days}. Together with Table~\ref{tab:model_performance}, they explain why a model ``performs well'' or ``fails'' in a domain: for example, GBoost's strong accuracy comes from low ATE but it fails stability checks across time groups (high WRS). These diagnostics support domain-aware model selection, focusing on the type of instability: series vs.\ month/day, that the ratings expose.

\section{Discussion and Conclusion}
This paper investigated three core research questions to evaluate the performance, interpretability, and robustness of forecasting models across diverse time-series domains. Our experiments and analyses yielded the following key insights:

\subsection {Summary of Findings}
\textbf{RQ1: When do forecasting models succeed or fail?}
Model performance is determined by the alignment between their inductive biases and the statistical properties of the data. GBoost demonstrated superior performance, achieving the lowest MASE across all domains by leveraging domain-specific feature engineering to handle challenges such as sparsity, non-stationarity, and irregular seasonality. In contrast, ARIMA performed competitively only on short, near-stationary series with strong autocorrelations, while Chronos excelled in finance but struggled in domains with extreme sparsity (e.g., car parts). The base Llama model showed degradation in forecasting accuracy, revealing that without domain-specific fine-tuning it cannot generalize to temporal patterns in the target data and is highly vulnerable to shifts between training and test distributions.

\noindent\textbf{RQ2: What do XAI methods reveal about model behavior?}
SHAP and LIME analyses provided mechanistic explanations for model successes and failures: GBoost relied heavily on engineered features (even if these features are as simple as domain-specific lags), confirming its adaptability to domain-specific temporal structures. Chronos's Surrogate revealed a tendency to default to statistical aggregates (e.g., expanding means) in complex domains, indicating its inability to leverage intricate patterns without explicit feature engineering. ARIMA’s assumptions, per LIME, led to over-reliance on recent lags, while LLaMA’s poor fine-tuning performance highlighted its dependence on pretraining data alignment.

\noindent\textbf{RQ3: How can rating metrics augment our understanding?}
The Average Treatment Effect (ATE) and Weighted Rejection Score (WRS) metrics quantified two critical dimensions of model reliability: Series-specific consistency (ATE): GBoost's low ATE in pedestrian and car datasets indicated uniform errors across series, while Chronos's low ATE in finance validated its generalization capability. Temporal stability (WRS): GBoost's high WRS revealed seasonal and day-of-week error variance, suggesting that its average accuracy may mask periodic unreliability. While our results highlight the promise of Rating-Driven Explanations (RDE), they are not directly compared against other fairness-oriented XAI approaches. Similarly, we did not evaluate RDE on synthetic benchmarks with ground-truth fairness properties. Both comparisons would provide a stronger test of our third research question and are left for future work. 

\subsection{Implications and Future Work}
Our findings show that effective forecasting depends on how well the modeling approach matches the structure and constraints of the domain. GBoost consistently outperforms other models when domain-specific features can be engineered (echoing the findings of \cite{classical-vs-dl-llm-ts}), demonstrating the strength of informed feature design over learned representations. Chronos and Llama-FT, while capable in domains with trend-driven data (e.g., finance), performed inconsistently in sparse or irregular settings (e.g., car sales). 
%For practitioners:1. Use GBoost when domain knowledge allows efficient feature engineering; 2. Apply time-series-specific FM like Chronos in domains with stable time-series patterns, but avoid them in sparse or highly irregular series; 3. Track ATE and WRS as complementary diagnostics: ATE for instability across series, WRS for instability across temporal groups. 
Significant opportunities exist to advance explainability in time-series forecasting. LIME adaptations could be more thoroughly explored, while this paper evaluates a baseline implementation (TS-MULE with uniform segmentation), prior work in classification has demonstrated more advanced techniques (e.g., dynamic segmentation or saliency-based sampling \cite{limesegment}) that may improve explanation stability for temporal data. Additionally, expanding the evaluation of forecast outputs beyond point predictions, such as analyzing the first predicted value, midpoint of the forecast horizon, or derived features (e.g., maximum/minimum values), could reveal new insights into model behavior across different temporal contexts. Third, incorporating physics-informed constraints into temporal foundation models (TFMs) may improve their ability to capture domain-specific dynamics while maintaining interpretability. Additionally, the proposed rating metrics (ATE/WRS) should be rigorously evaluated as XAI methods themselves, following frameworks like \cite{MERSHA2024128111} to quantify their effectiveness in surfacing model biases. Finally, combining these directions could yield unified evaluation protocols that assess models holistically across accuracy, robustness, and explainability dimensions.

\bigskip

\bibliography{references}
\clearpage
\appendix
\newpage
\setcounter{section}{0}
\renewcommand{\thesection}{\Alph{section}}
\setcounter{figure}{0}
\renewcommand{\thefigure}{\thesection.\arabic{figure}}
\setcounter{table}{0}
\renewcommand{\thetable}{\thesection.\arabic{table}}

\section*{Appendix}

\section{RDE}
\label{app:rde}
\begin{figure}[h]
    \centering
\includegraphics[width=0.45\textwidth]{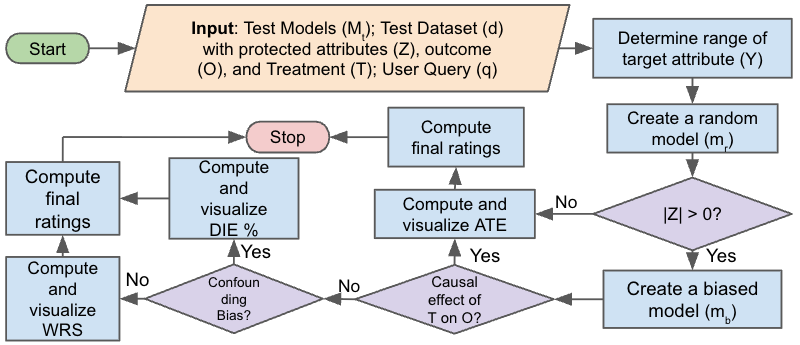}
    \caption{RDE Workflow.}
    \label{fig:rde-workflow}
\end{figure}

Although \cite{lakkaraju2025h-xai} also considered another rating metric, DIE \%, we restrict our analysis to two metrics, ATE and WRS, to validate the hypotheses in Tables \ref{tab:ate} and \ref{tab:wrs}:

\noindent \textbf{Weighted Rejection Score (WRS):} WRS quantifies statistical bias across a protected attribute \( z_i \in Z \) by comparing outcome distributions \( (O \mid z_i) \) across all \( \binom{m}{2} \) group pairs using Student’s t-test~\cite{student1908probable}. For each pair, the t-value \( t_{z_i} \) is compared to the critical value \( t_{\text{crit}} \) for a given confidence interval (CI). A rejection indicates a significant difference in outcomes. WRS is computed as a weighted sum of rejections across CIs, where \( x_i \) is the number of rejections and \( w_i \) is the assigned weight (e.g., 1, 0.8, 0.6 for 95\%, 75\%, and 60\% CIs). Formally, $WRS = \sum_{i} w_i \cdot x_i$.

\noindent \textbf{Average Treatment Effect (ATE):} Average Treatment Effect (ATE) \cite{rubin1974estimating,splawa1990application} measures the average difference in outcomes between the treated ($T = i$) and untreated ($T = 0$) units, quantifying the impact of the treatment on the AI model's outcome. Unlike local counterfactuals, ATE quantifies the average effect of a treatment on model predictions across the dataset, making it well-suited for answering stakeholder questions about overall effect of an attribute on the outcome. Formally, $ATE  = [|E[O| do(T = i)] - E[O| do(T = 0)]| ]$

\section{SHAP Analysis}
\label{app:shap}

\begin{figure}[htbp]
    \centering
    \begin{subfigure}[b]{0.23\textwidth}
        \centering        \includegraphics[width=\linewidth]{figures/SHAP/Chronos/shap_summary_finance.png}
        \caption{Finance}
        \label{fig:chronos_shap_finance}
    \end{subfigure}
    \hfill
    \begin{subfigure}[b]{0.23\textwidth}
        \centering
        \includegraphics[width=\linewidth]{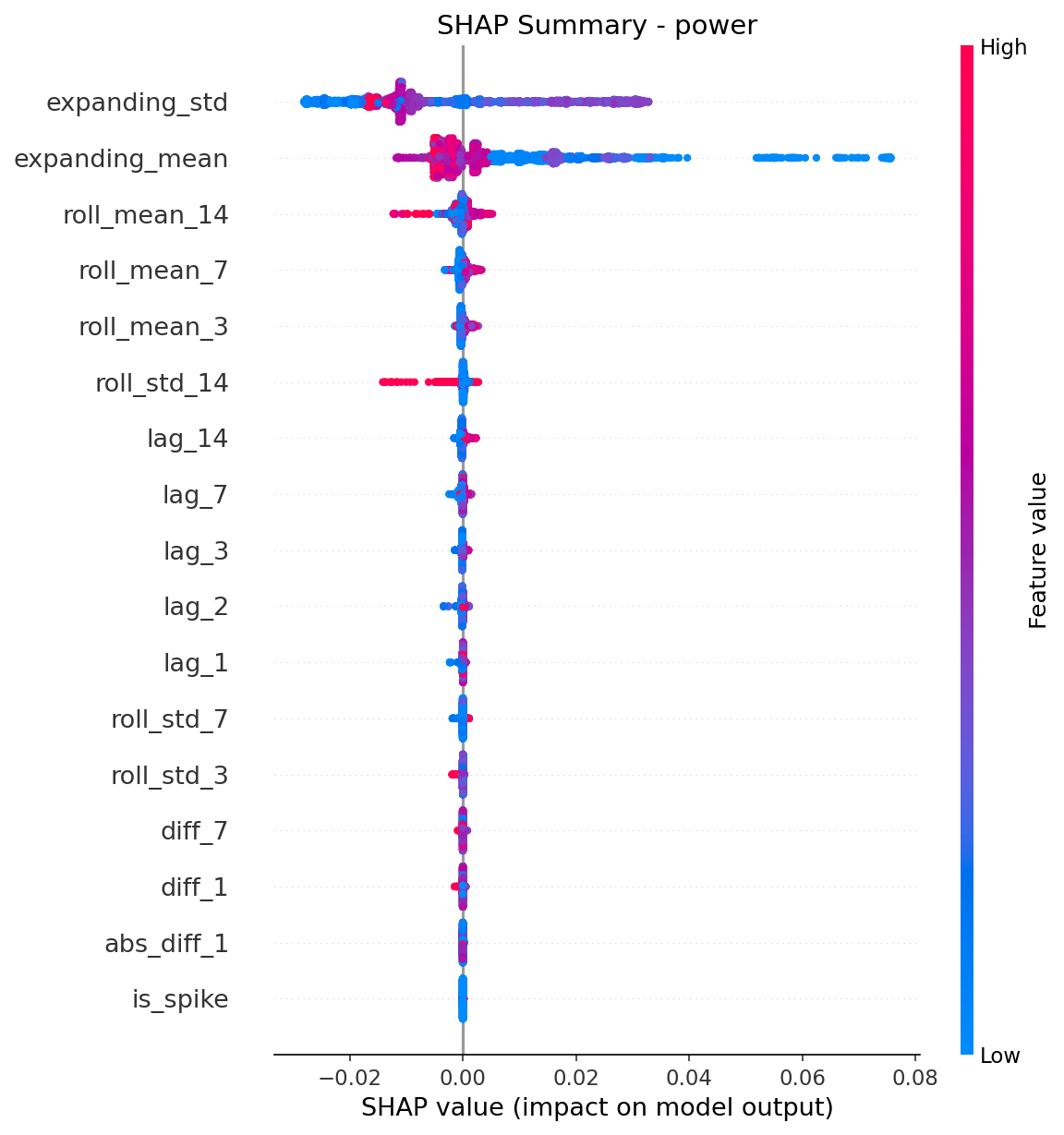}
        \caption{Power}
        \label{fig:chronos_shap_power}
    \end{subfigure}
    \hfill
    \begin{subfigure}[b]{0.23\textwidth}
        \centering
        \includegraphics[width=\linewidth]{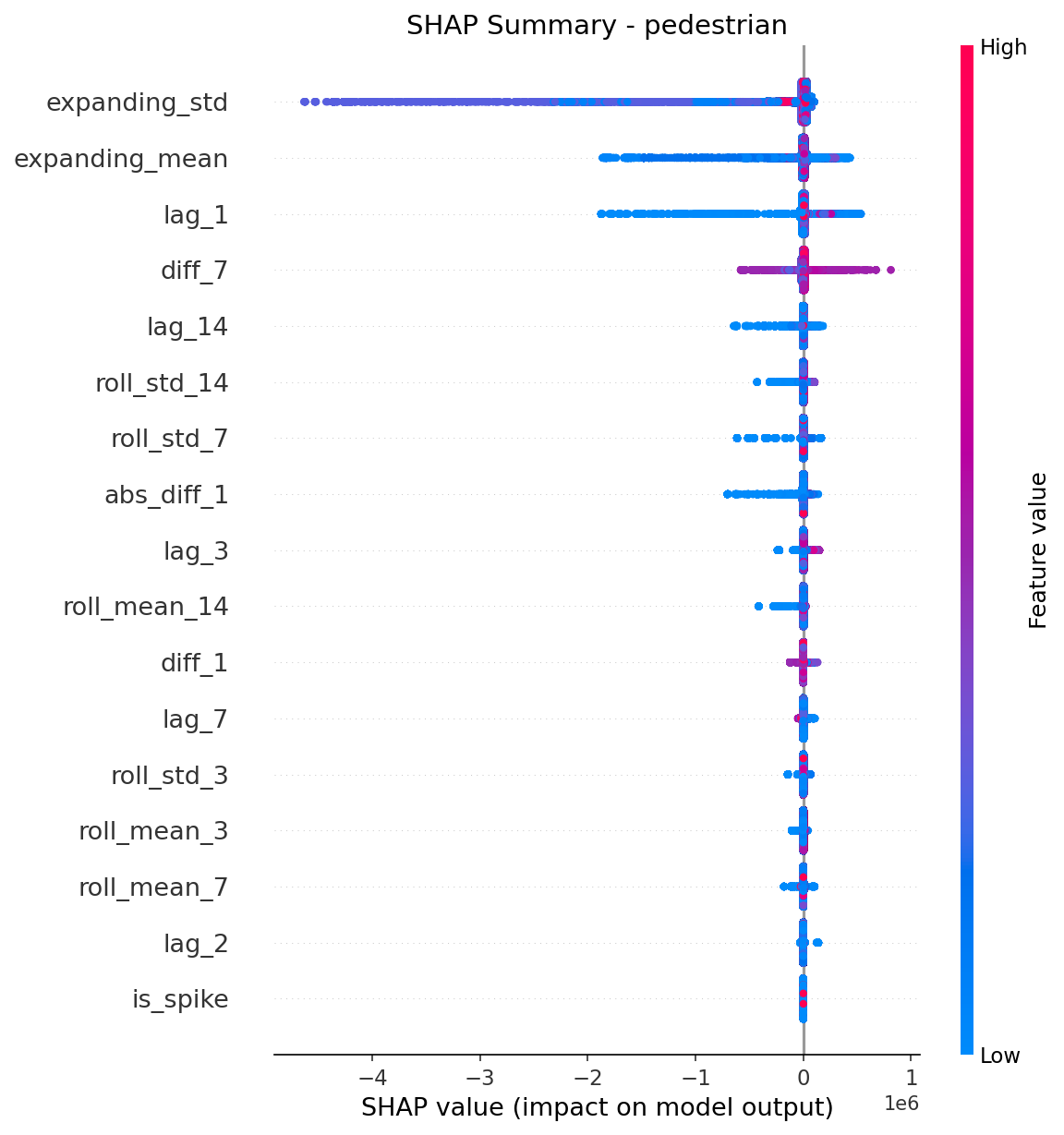}
        \caption{Pedestrian}
        \label{fig:chronos_shap_pedestrian}
    \end{subfigure}
    \hfill
    \begin{subfigure}[b]{0.23\textwidth}
        \centering
        \includegraphics[width=\linewidth]{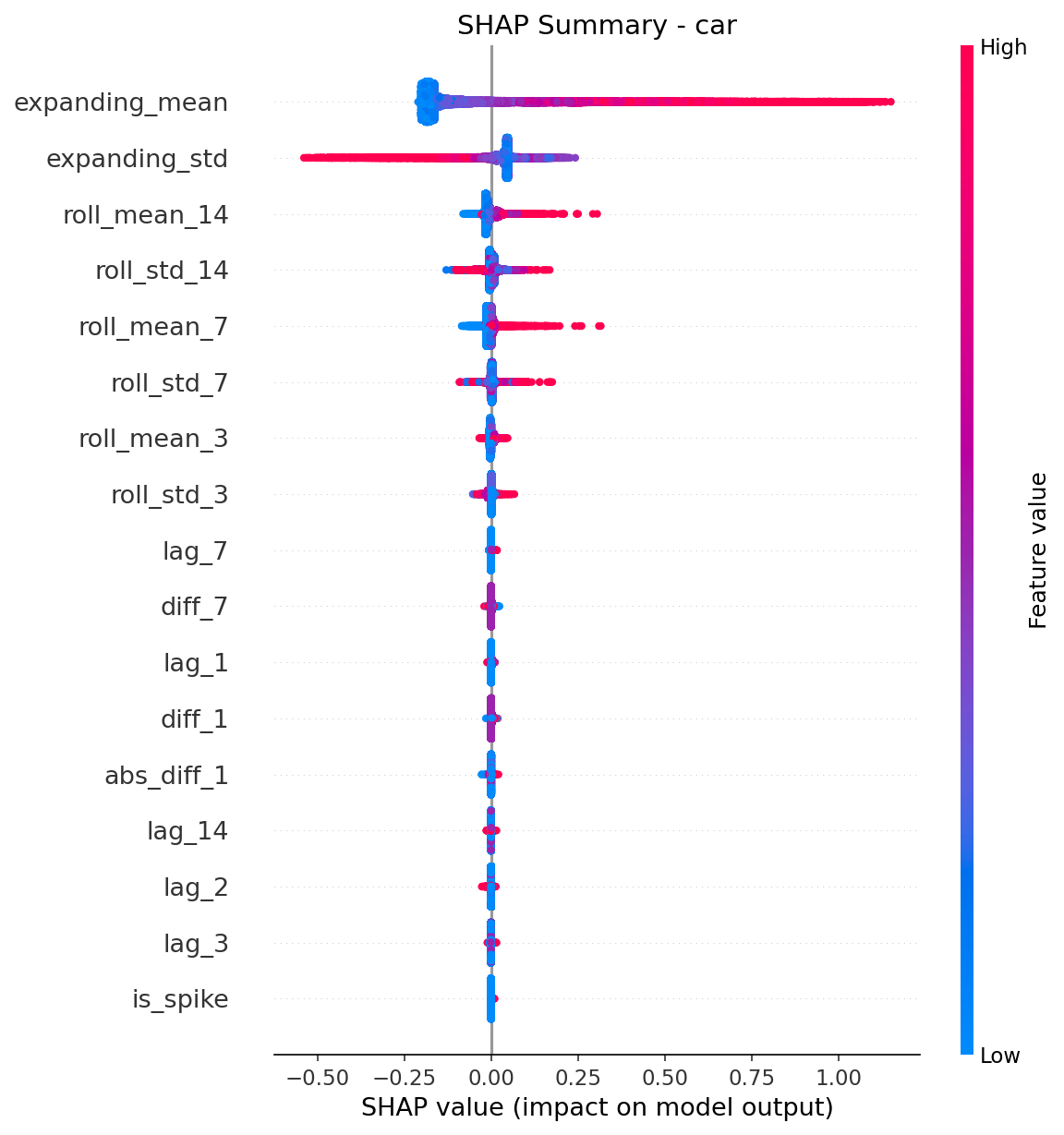}
        \caption{Car}
        \label{fig:chronos_shap_car}
    \end{subfigure}
    \caption{SHAP values for Chronos Surrogate model across different domains.}
    \label{fig:chronos_shap_values}
\end{figure}
In \textbf{Power}, statistical features outperform lags: all rolling means rank above any lag feature. Expanding standard deviation has a strong negative impact - high volatility triggers downward predictions. Power is a volatile domain where ``default to mean'' becomes dominant strategy. From Figure \ref{fig:chronos_shap_values}, we see that in the \textbf{Pedestrian} domain, lags make up 2 of the top 5 features. Once again, expanding std. emerges as the most important. The other important statistical features were expanding mean and the difference between the last time pt and lag 7. Rolling means prove irrelevant. Given the size of the pedestrian dataset and the lack of feature engineering for the surrogate model it is obvious that the short lags fail to capture the daily and weekly seasonality. In the \textbf{Car} domain, expanding mean dominates completely - model collapses to baseline in sparse domain. High feature values correlate linearly with SHAP values, indicating limited pattern capture. Explains discrepancy: decent SMAPE (mean-relative) but catastrophic MASE (vs naive). 

\begin{figure}[htbp]
    \centering
    \begin{subfigure}[b]{0.23\textwidth}
        \centering
        \includegraphics[width=\linewidth]{figures/SHAP/GB/GB_SHAP_finance.png}
        \caption{Finance}
        \label{fig:gb_shap_finance}
    \end{subfigure}
    \hfill
    \begin{subfigure}[b]{0.23\textwidth}
        \centering
        \includegraphics[width=\linewidth]{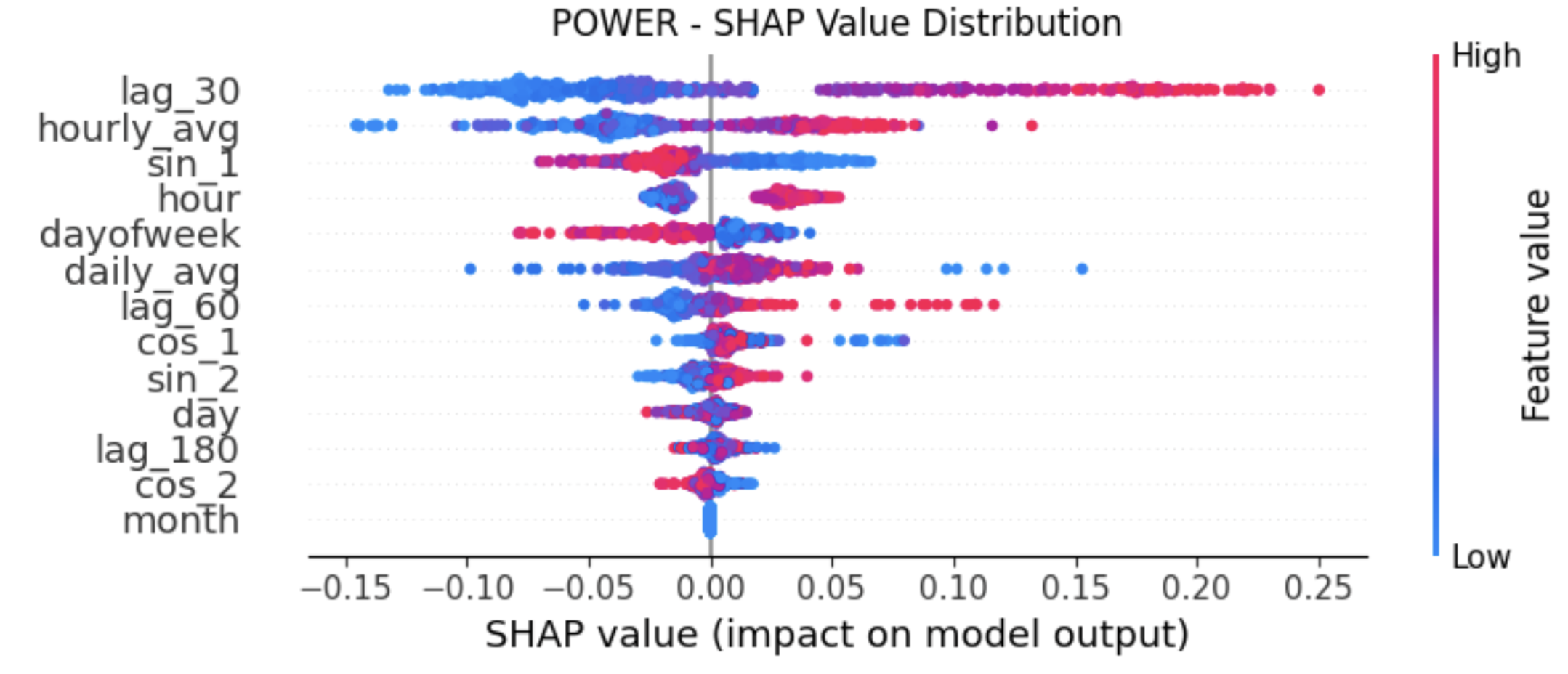}
        \caption{Power}
        \label{fig:gb_shap_power}
    \end{subfigure}
    \hfill
    \begin{subfigure}[b]{0.23\textwidth}
        \centering
        \includegraphics[width=\linewidth]{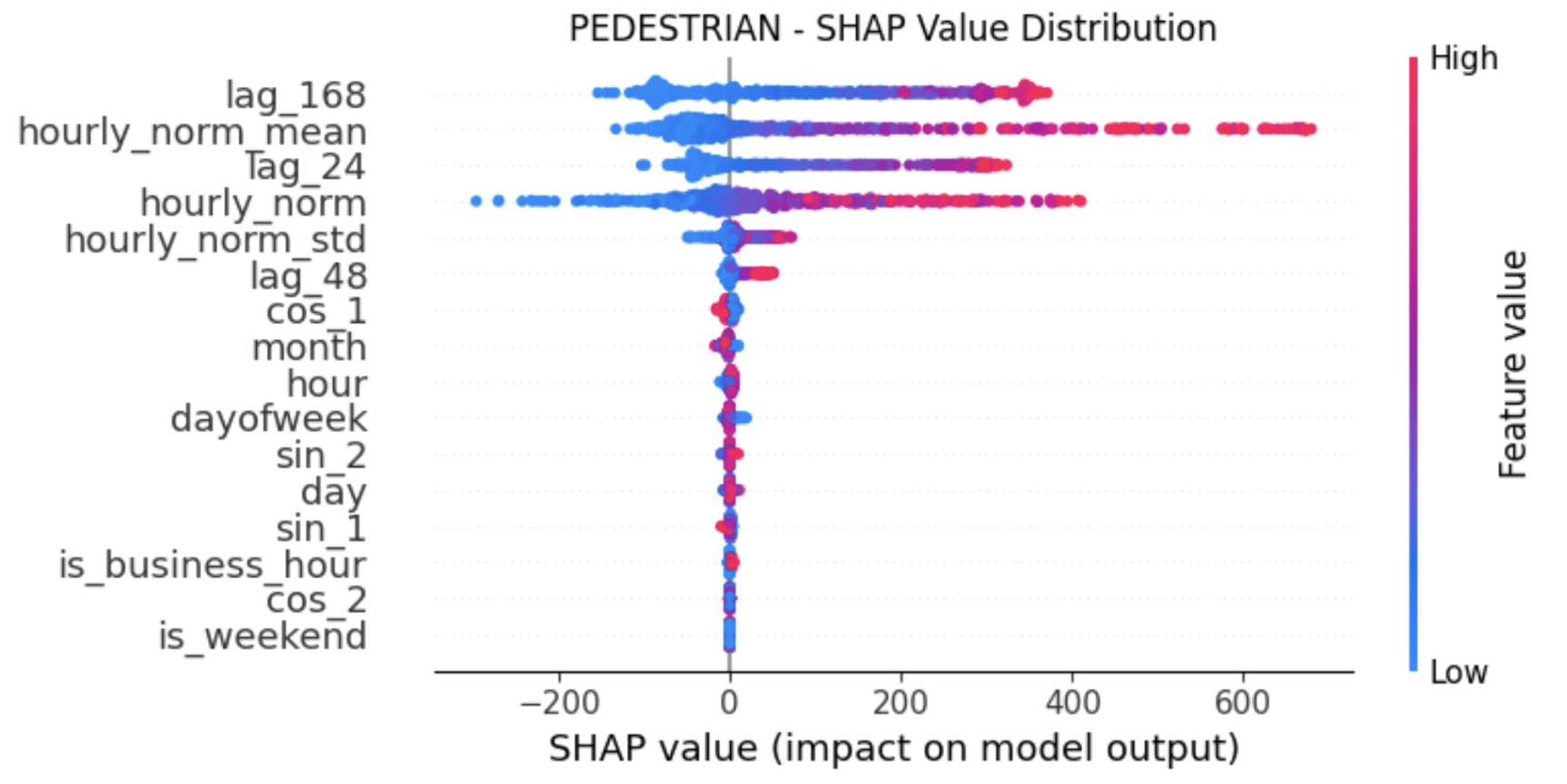}
        \caption{Pedestrian}
        \label{fig:gb_shap_pedestrian}
    \end{subfigure}
    \hfill
    \begin{subfigure}[b]{0.23\textwidth}
        \centering
        \includegraphics[width=\linewidth]{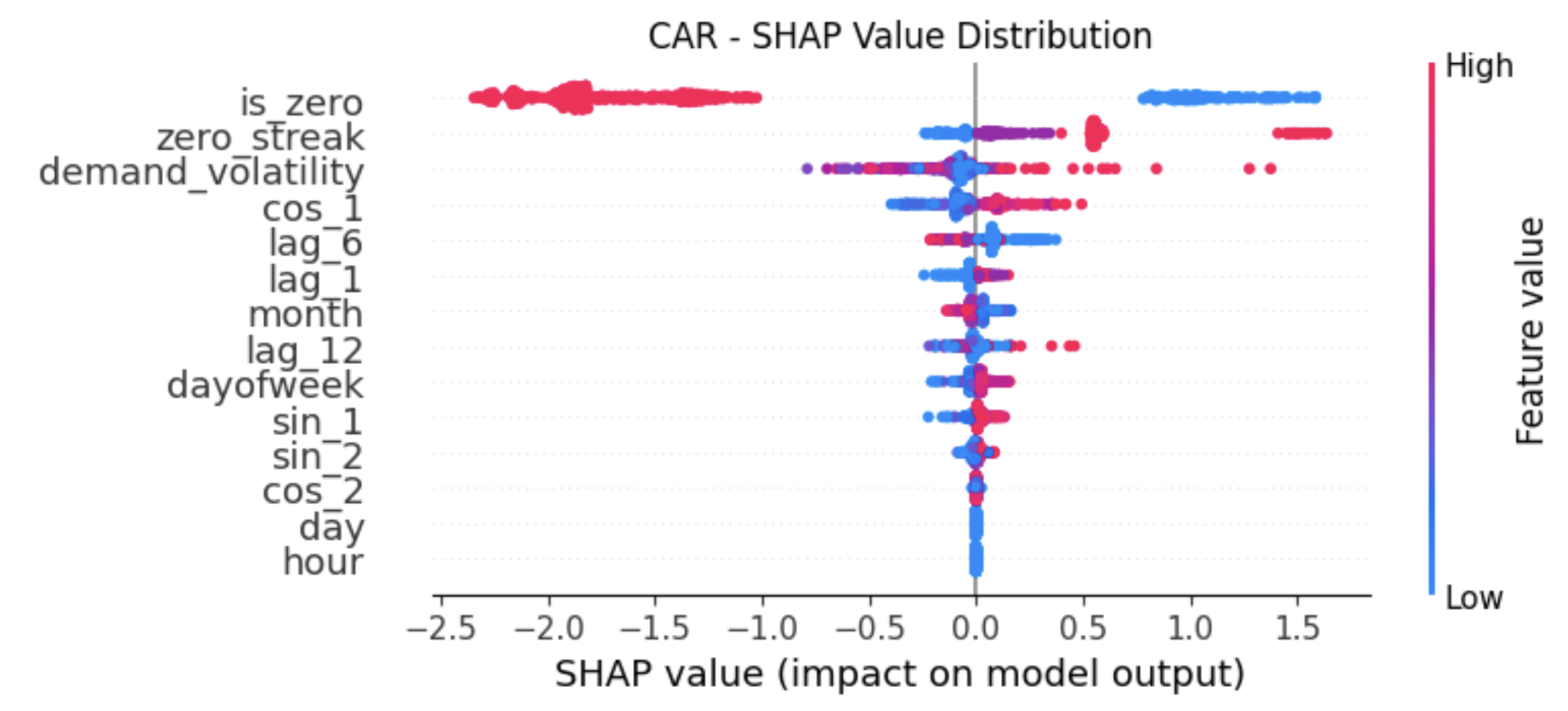}
        \caption{Car}
        \label{fig:gb_shap_car}
    \end{subfigure}
    \caption{SHAP values for Gradient Boosting model across different domains.}
    \label{fig:gb_shap_values}
\end{figure}
For \textbf{Power}, lag-30 (30-minute intervals) emerged as most critical, likely reflecting meeting schedules in the observed building. Hourly averages ranked second, capturing behavioral routines. The feature importance profile mirrors human activity patterns: arrivals, lunch breaks, and departures create predictable power demand fluctuations. %Gradient Boosting's superior performance (only model beating naive baseline) stems from its ability to model these fine-grained temporal dependencies. 
With regards to \textbf{Pedestrian}, lag-168 (weekly seasonality) and Lag-24 (daily cycle) dominated, reflecting strong periodicity in foot traffic. Surprisingly, a binary weekend indicator showed minimal impact - likely because weekly lags already encode this structural information. Sensor location heterogeneity (business vs. leisure zones) may explain why simple temporal markers underperformed compared to lagged values.
In the final domain,\textbf{Car}, the most important feature was whether the previous observation was zero, highlighting the dataset's extreme sparsity. Engineered features outperformed raw lags, demonstrating the need for domain-specific representations in sparse, spike-prone environments. All models failed to achieve satisfactory performance (MASE $>$ 1.0), underscoring the unique challenges of zero-inflated time-series.

\begin{table}
    \centering
    \small  % Slightly smaller font for better fit
    \begin{tabular}{l l r@{\,±\,}l r@{\,±\,}l r@{\,±\,}l r@{\,±\,}l}
        \toprule 
        \textbf{Domain} & \textbf{Metric} & \multicolumn{2}{c}{\textbf{Chronos}} & \multicolumn{2}{c}{\textbf{Chronos-S}}  \\
        \midrule
        \textbf{Finance} & MASE & 5.48 & 5.44 & 5.60 & 5.34 \\
                         & SMAPE & 5.30 & 5.15 & 5.40 & 5.05  \\
        \midrule
        \textbf{Power} & MASE & 2.33 & 0.00 & 1.29 & 0.00 \\
                      & SMAPE & 8.84\% & 0.00\% & 5.02 & 0.00\% \\
        \midrule
        \textbf{Pedestrian} & MASE & 1.47 & 0.67* & 0.71 & 0.24\\
                           & SMAPE 
                           & 90.98 & 44.77*  & 38.88 & 13.29 \\
        \midrule    
        \textbf{Car Parts} & MASE & \multicolumn{2}{c}{$\infty$} & 0.79 & 1.32 \\
                          & SMAPE & 175.03 & 37.56* & 190.49 & 16.30  \\
        \bottomrule
    \end{tabular}
    \caption{Forecasting performance comparison. MASE and sMAPE are reported as (mean $\pm$ std. dev.). sMAPE is reported as percentage (\%). Chronos Surrogate $\rightarrow$ Chronos-S. Chronos-S performs comparably to or better than Chronos without domain specific feature engineering. 
    * Chronos has already seen these datasets in training and our results may differ from the authors' due to different experimental setups. } 
    
    \label{tab:surrogate_performance}
\end{table}

\section{LIME Analysis}
\label{app:lime}

\begin{figure*}[htbp]
    \centering
    \begin{subfigure}[b]{0.45\textwidth}
        \centering
        \includegraphics[width=\linewidth]{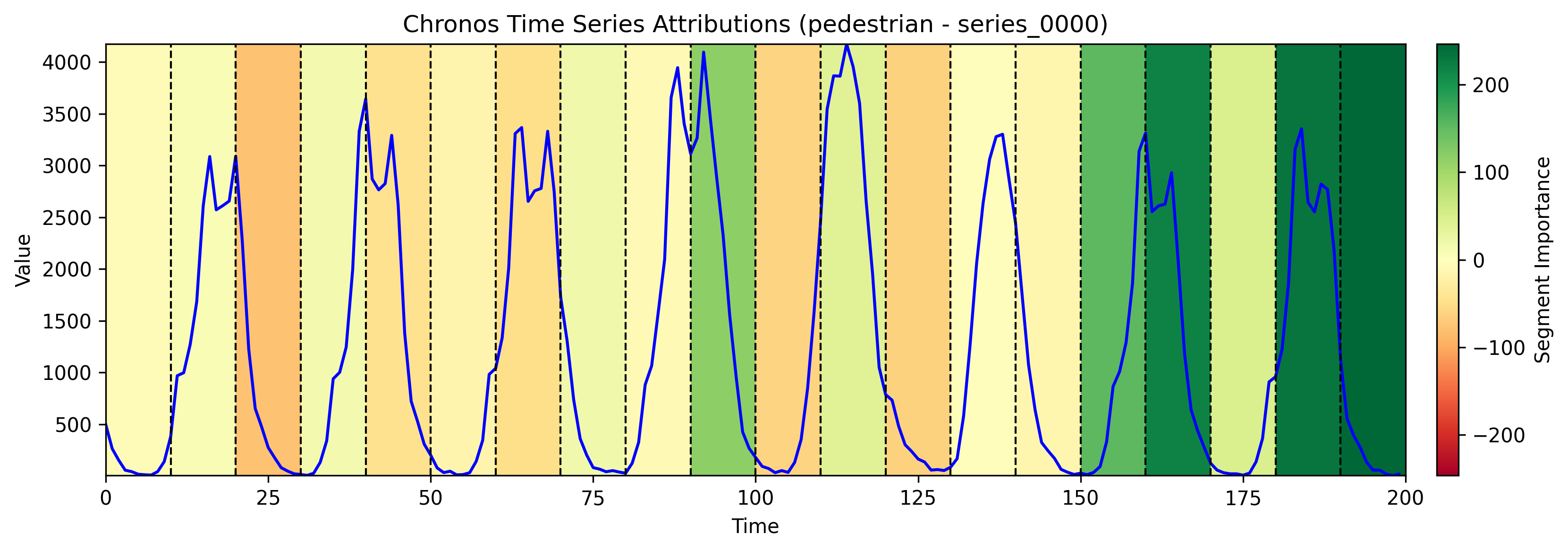}
        \caption{Pedestrian series\_0}
        \label{fig:Chronos_LIME_pedestrian0}
    \end{subfigure}
    \hfill
    \begin{subfigure}[b]{0.45\textwidth}
        \centering
        \includegraphics[width=\linewidth]{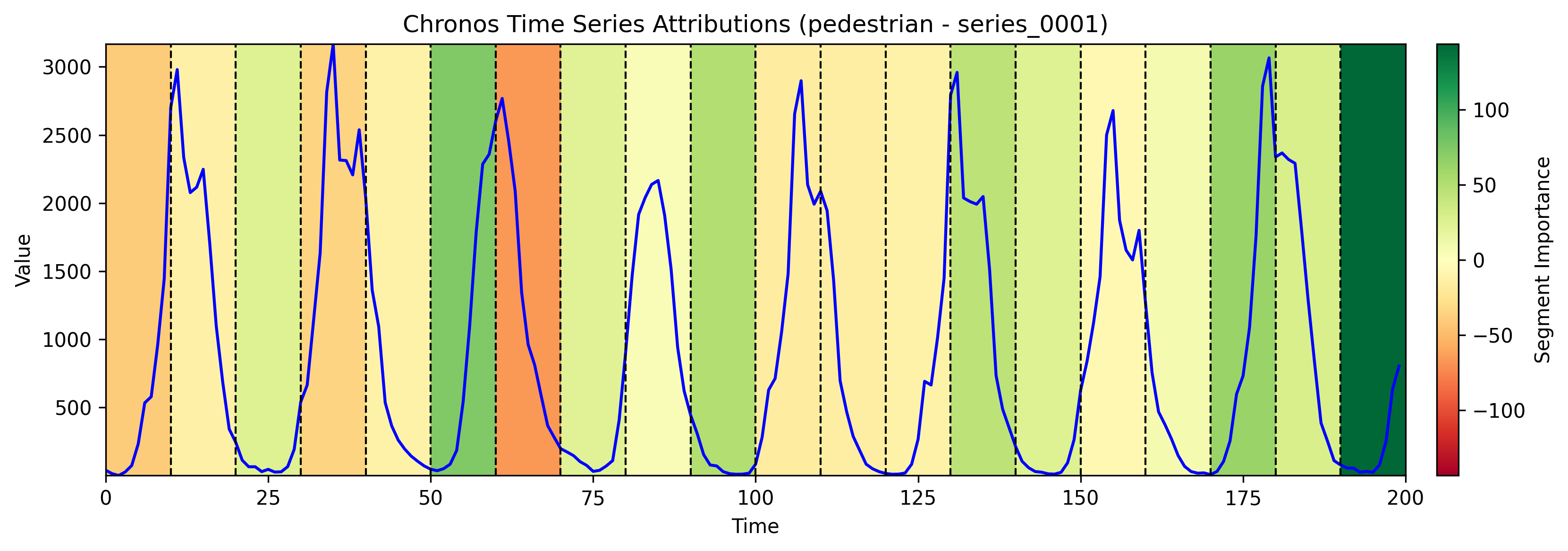}
        \caption{Pedestrian series\_1}
        \label{fig:Chronos_LIME_pedestrian1}
    \end{subfigure}
    \hfill
    \begin{subfigure}[b]{0.45\textwidth}
        \centering
        \includegraphics[width=\linewidth]{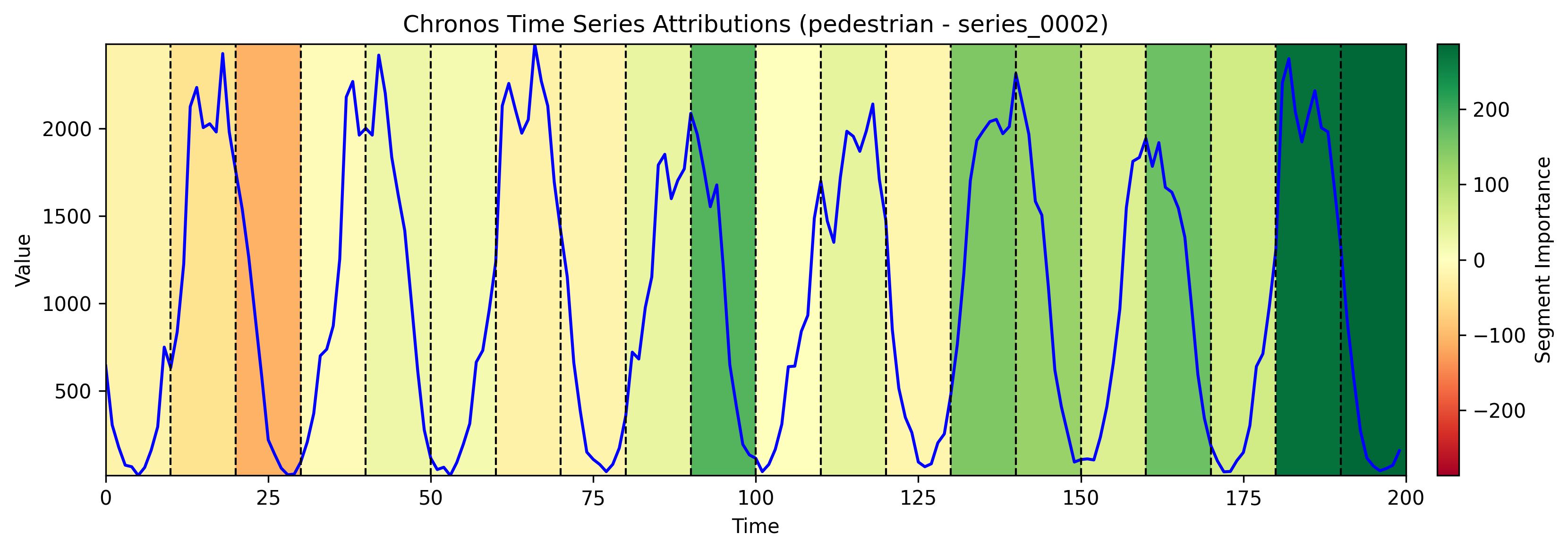}
        \caption{Pedestrian series\_2}
        \label{fig:Chronos_LIME_pedestrian2}
    \end{subfigure}
    \hfill
    \begin{subfigure}[b]{0.45\textwidth}
        \centering
        \includegraphics[width=\linewidth]{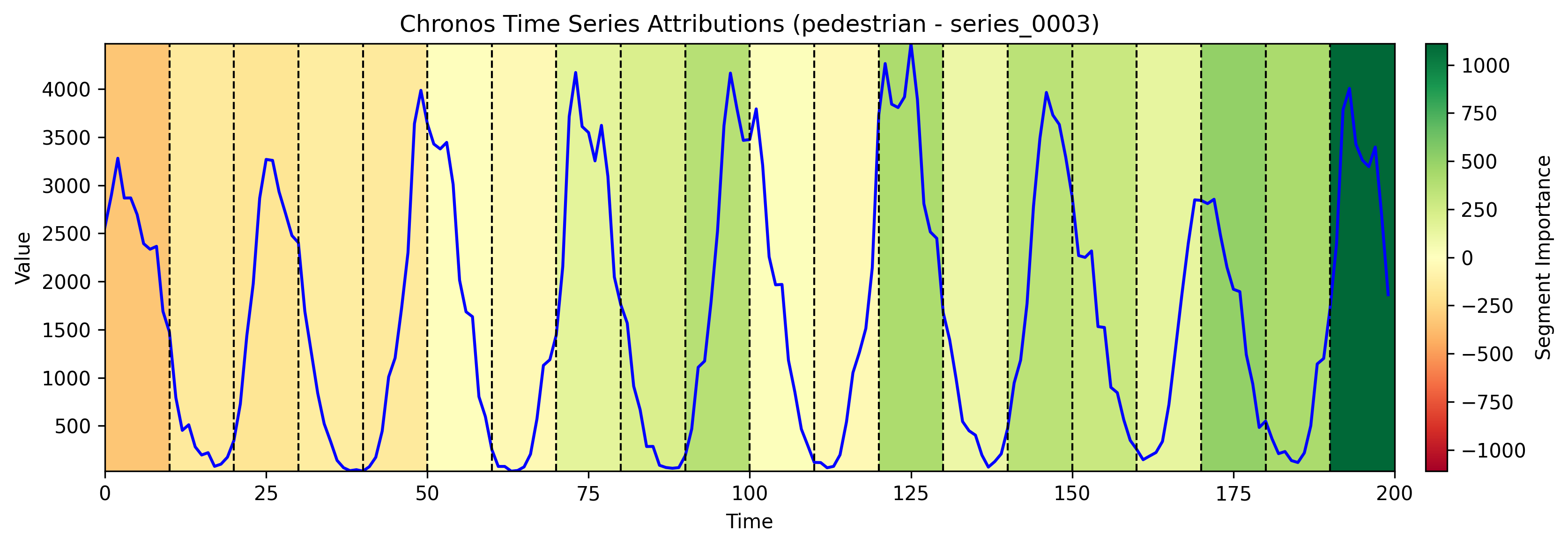}
        \caption{Pedestrian series\_3}
        \label{fig:Chronos_LIME_pedestrian3}
    \end{subfigure}
    \hfill
    \begin{subfigure}[b]{0.45\textwidth}
        \centering
        \includegraphics[width=\linewidth]{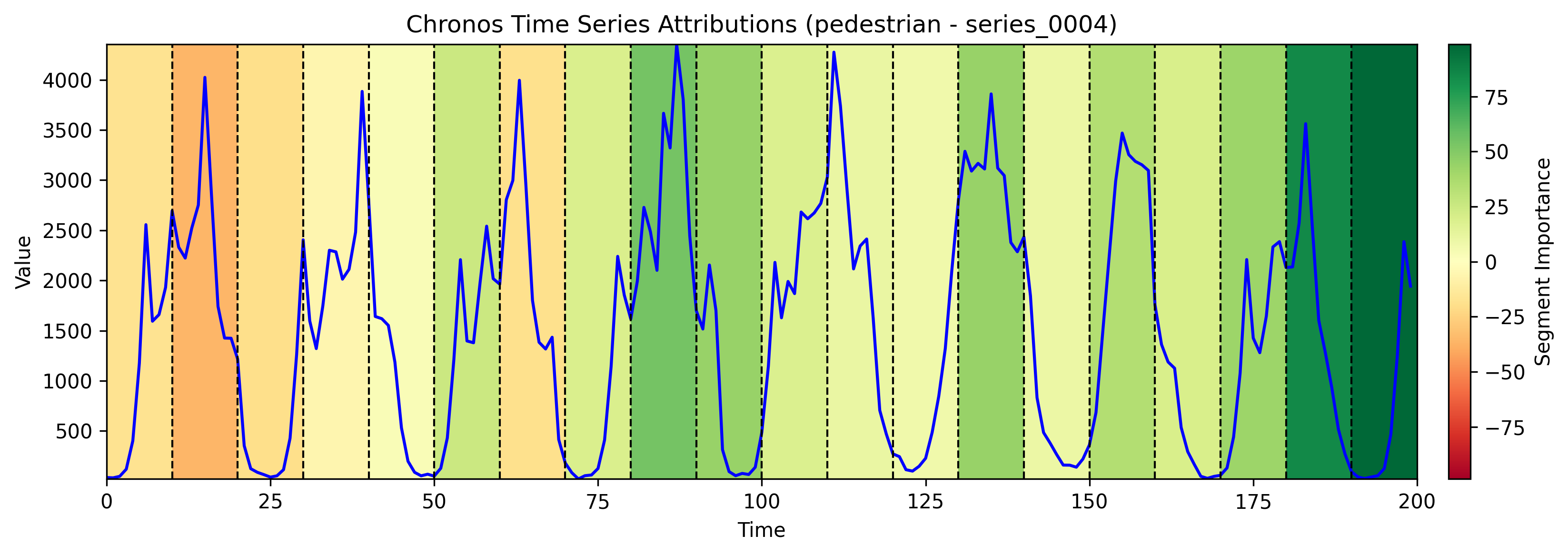}
        \caption{Pedestrian series\_4}
        \label{fig:Chronos_LIME_pedestrian4}
    \end{subfigure}
    \hfill
    \begin{subfigure}[b]{0.45\textwidth}
        \centering
        \includegraphics[width=\linewidth]{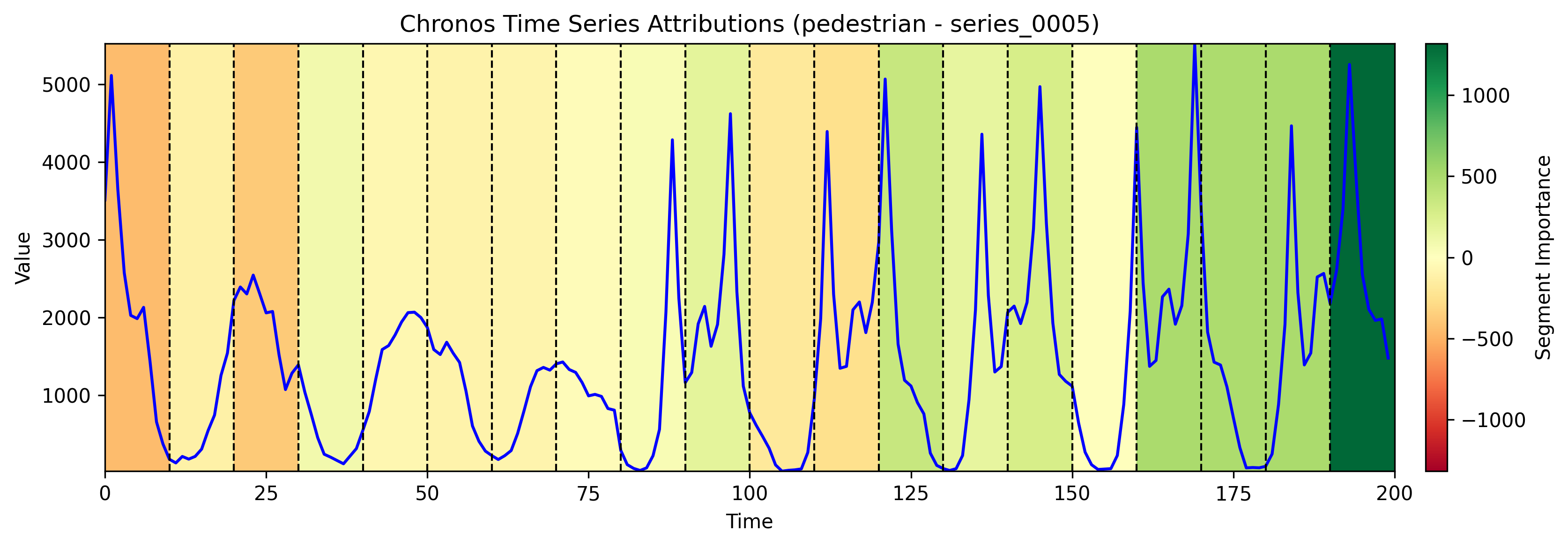}
        \caption{Pedestrian series\_5}
        \label{fig:Chronos_LIME_pedestrian5}
    \end{subfigure}
    \hfill
    \begin{subfigure}[b]{0.45\textwidth}
        \centering
        \includegraphics[width=\linewidth]{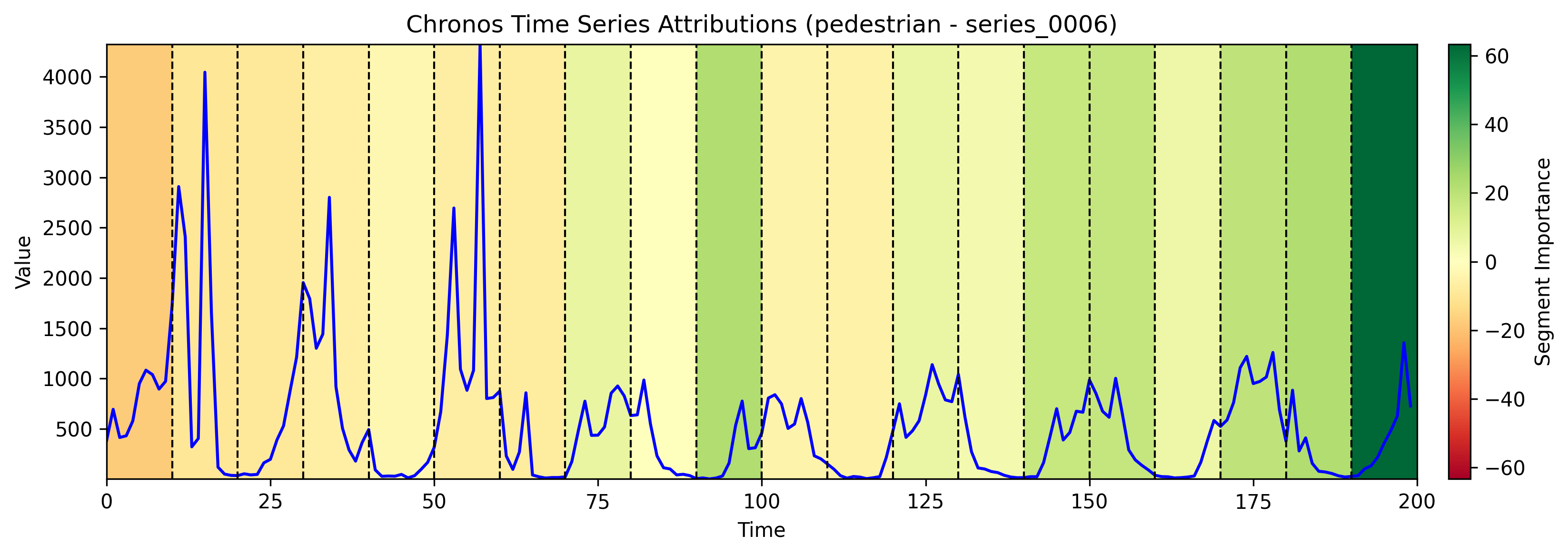}
        \caption{Pedestrian series\_6}
        \label{fig:Chronos_LIME_pedestrian6}
    \end{subfigure}
    \hfill
    \begin{subfigure}[b]{0.45\textwidth}
        \centering
        \includegraphics[width=\linewidth]{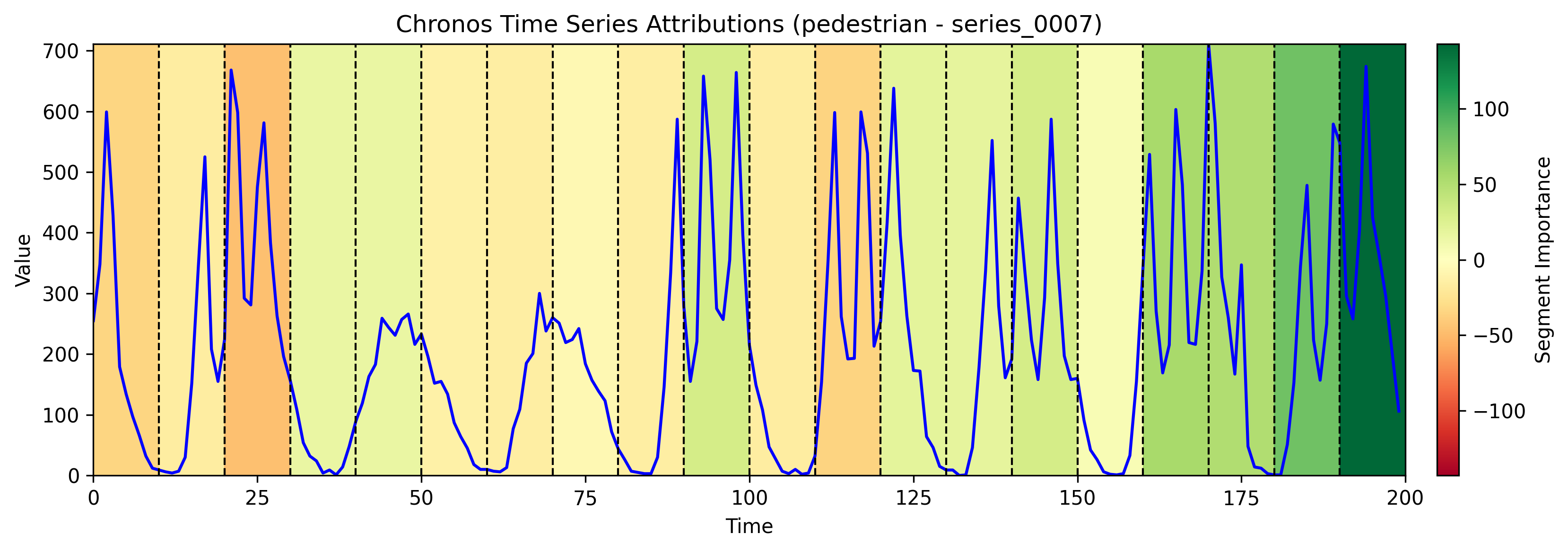}
        \caption{Pedestrian series\_7}
        \label{fig:Chronos_LIME_pedestrian7}
    \end{subfigure}
    \hfill
    \begin{subfigure}[b]{0.45\textwidth}
        \centering
        \includegraphics[width=\linewidth]{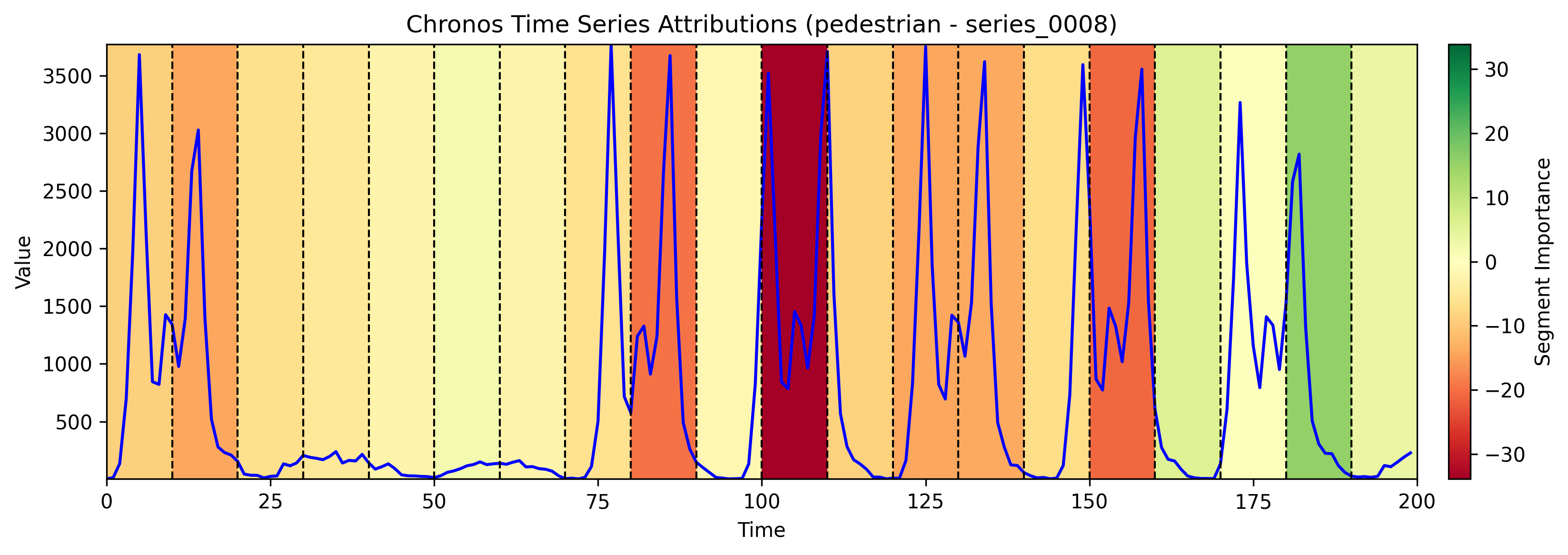}
        \caption{Pedestrian series\_8}
        \label{fig:Chronos_LIME_pedestrian8}
    \end{subfigure}
    \hfill
    \begin{subfigure}[b]{0.45\textwidth}
        \centering
        \includegraphics[width=\linewidth]{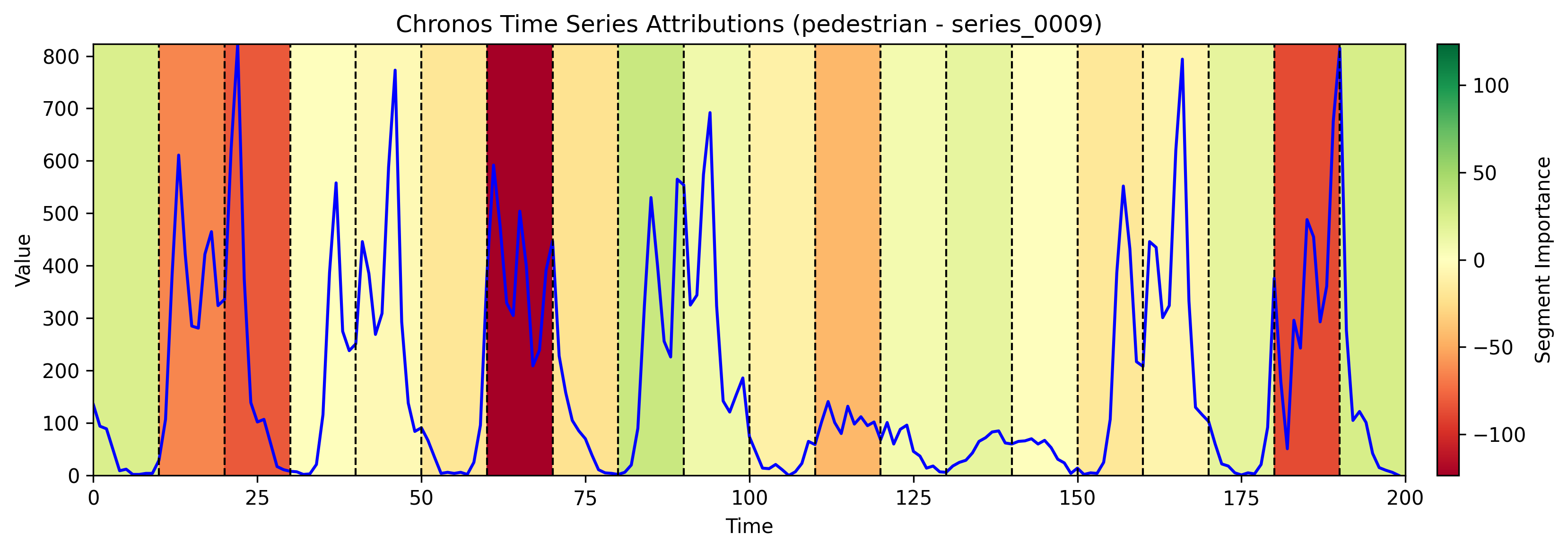}
        \caption{Pedestrian series\_9}
        \label{fig:Chronos_LIME_pedestrian9}
    \end{subfigure}
    \hfill
    \begin{subfigure}[b]{0.45\textwidth}
        \centering
        \includegraphics[width=\linewidth]{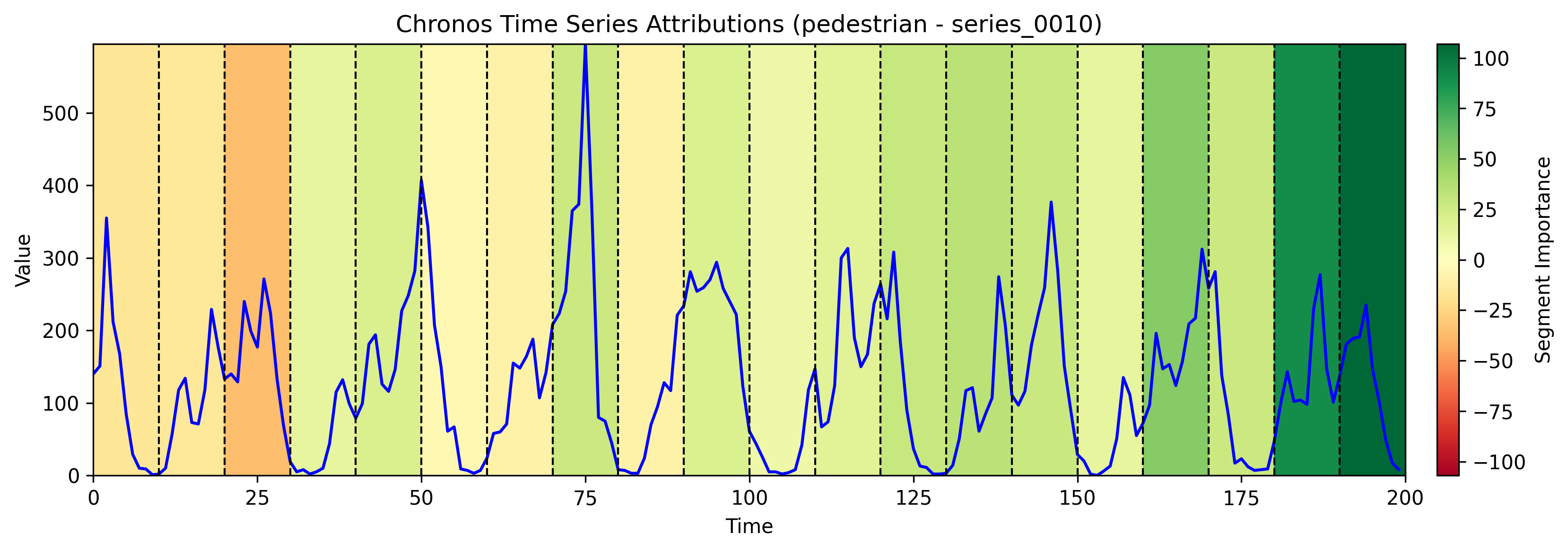}
        \caption{Pedestrian series\_10}
        \label{fig:Chronos_LIME_pedestrian10}
    \end{subfigure}
    \hfill
    \begin{subfigure}[b]{0.45\textwidth}
        \centering
        \includegraphics[width=\linewidth]{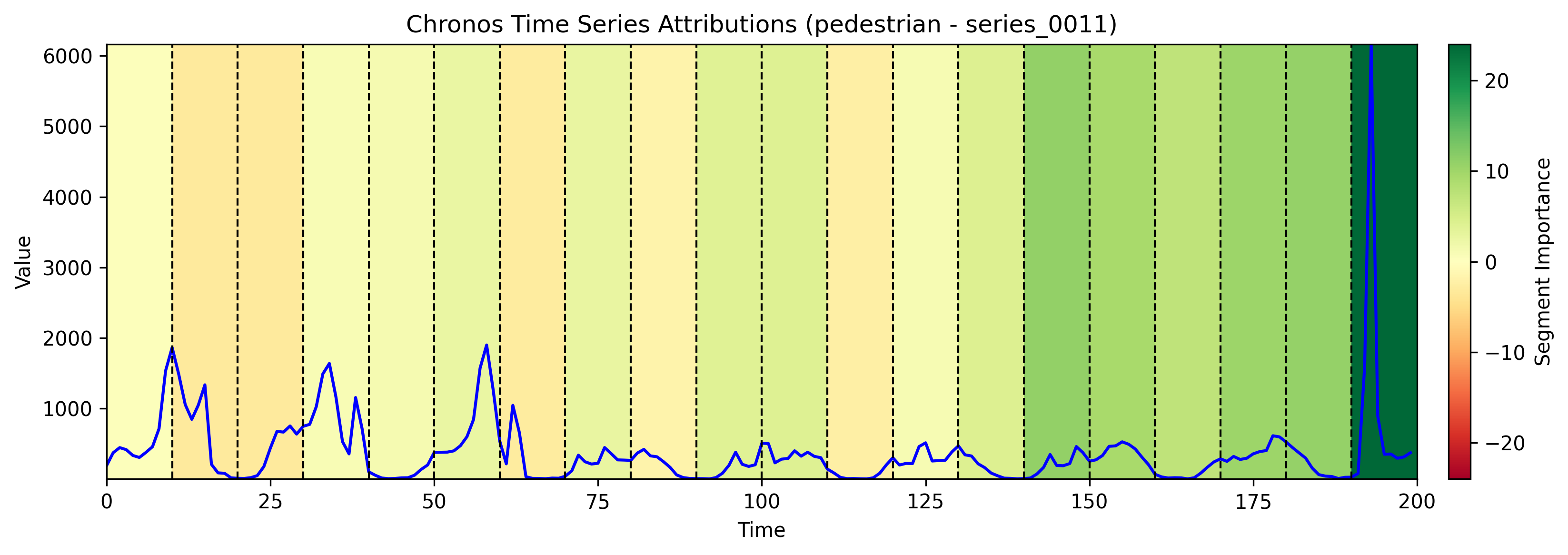}
        \caption{Pedestrian series\_11}
        \label{fig:Chronos_LIME_pedestrian11}
    \end{subfigure}
    
    \caption{LIME plots for Chronos model in the pedestrian domain using 20 uniform segments, zero replacement, and 100 samples on the last 200 points of the input time-series. Figures show the distribution of feature importance for each time-series, revealing a bias for negative feature importance in earlier segments and positive feature importance in later segments.}
    \label{fig:chronos_lime_pedestrian_values}
\end{figure*}

\begin{figure*}[htbp]
    \centering
    \begin{subfigure}[b]{0.45\textwidth}
        \centering
        \includegraphics[width=\linewidth]{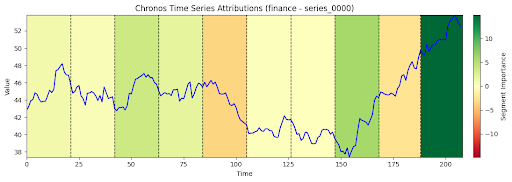}
        \caption{Finance series\_0}
        \label{fig:Chronos_LIME_finance0_10seg}
    \end{subfigure}
    \hfill
    \begin{subfigure}[b]{0.45\textwidth}
        \centering
        \includegraphics[width=\linewidth]{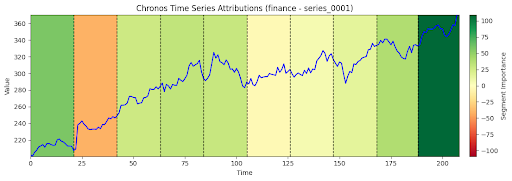}
        \caption{Finance series\_1}
        \label{fig:Chronos_LIME_finance1_10seg}
    \end{subfigure}
    \hfill
    \begin{subfigure}[b]{0.45\textwidth}
        \centering
        \includegraphics[width=\linewidth]{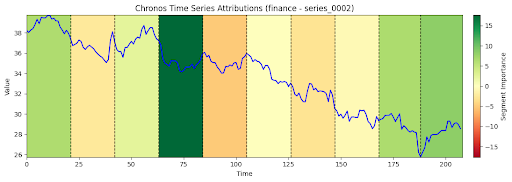}
        \caption{Finance series\_2}
        \label{fig:Chronos_LIME_finance2_10seg}
    \end{subfigure}
    \hfill
    \begin{subfigure}[b]{0.45\textwidth}
        \centering
        \includegraphics[width=\linewidth]{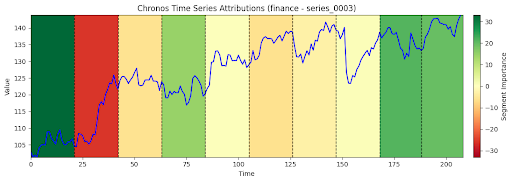}
        \caption{Finance series\_3}
        \label{fig:Chronos_LIME_finance3_10seg}
    \end{subfigure}
    \hfill
    \begin{subfigure}[b]{0.45\textwidth}
        \centering
        \includegraphics[width=\linewidth]{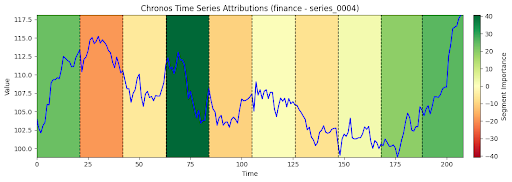}
        \caption{Finance series\_4}
        \label{fig:Chronos_LIME_finance4_10seg}
    \end{subfigure}
    \hfill
    \begin{subfigure}[b]{0.45\textwidth}
        \centering
        \includegraphics[width=\linewidth]{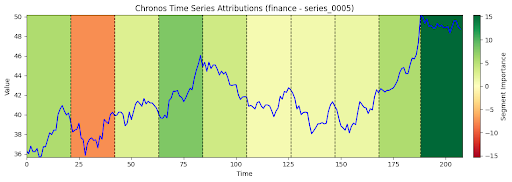}
        \caption{Finance series\_5}
        \label{fig:Chronos_LIME_finance5_10seg}
    \end{subfigure}
    \caption{LIME plots for Chronos model in finance using 10 uniform segments, zero replacement, and 200 samples.}
    \label{fig:chronos_lime_values_finance_10seg}
\end{figure*}

\begin{figure*}[htbp]
    \centering
    \begin{subfigure}[b]{0.45\textwidth}
        \centering
        \includegraphics[width=\linewidth]{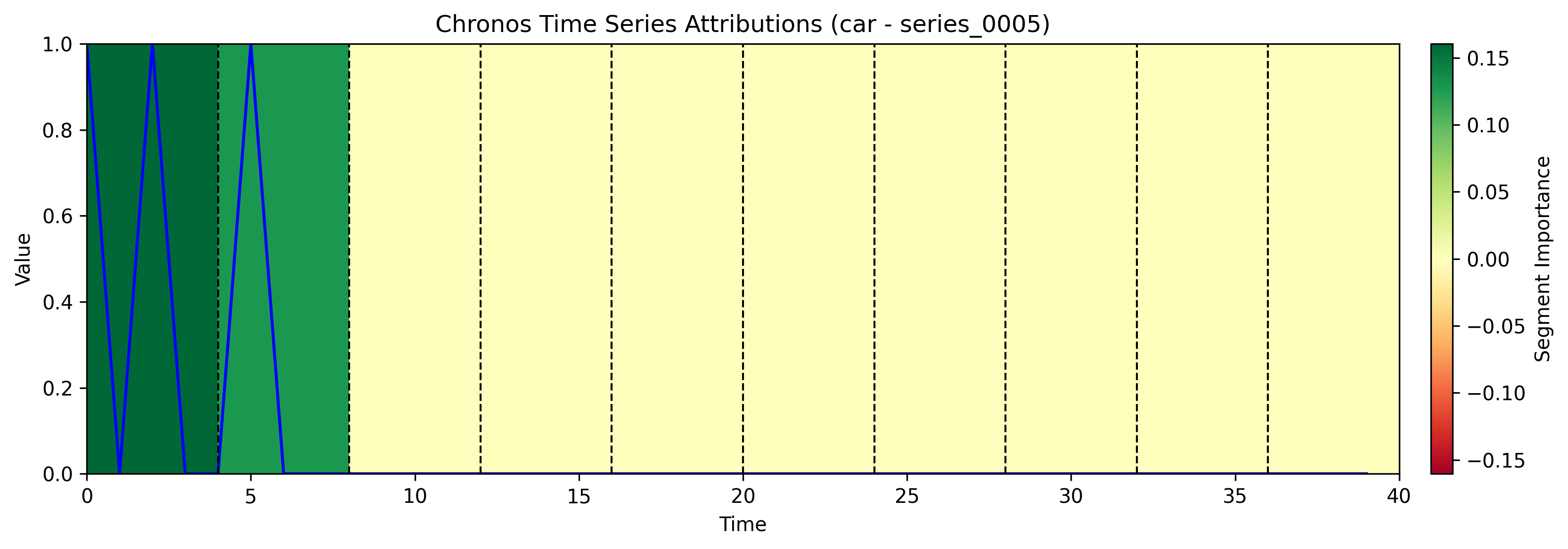}
        \caption{Car series\_5}
        \label{fig:Chronos_LIME_car5_zero}
    \end{subfigure}
    \hfill
    \begin{subfigure}[b]{0.45\textwidth}
        \centering
        \includegraphics[width=\linewidth]{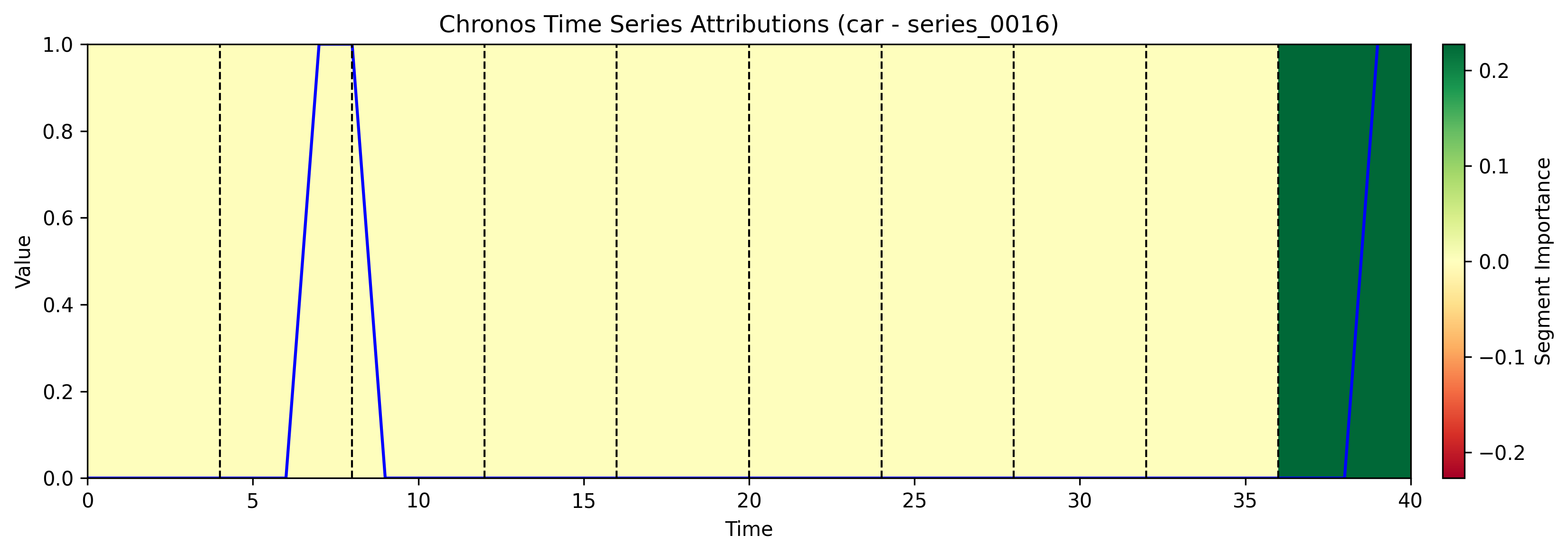}
        \caption{Car series\_16}
        \label{fig:Chronos_LIME_car16_zero}
    \end{subfigure}
    \hfill
    \begin{subfigure}[b]{0.45\textwidth}
        \centering
        \includegraphics[width=\linewidth]{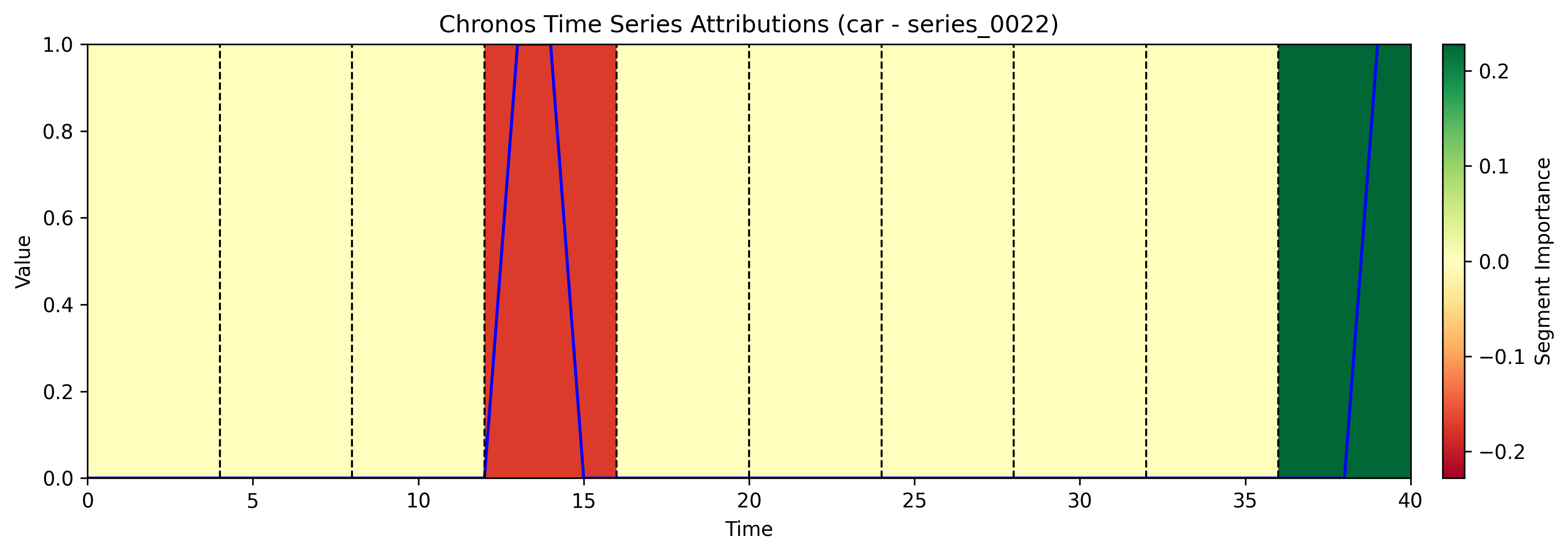}
        \caption{Car series\_22}
        \label{fig:Chronos_LIME_car22_zero}
    \end{subfigure}
    \hfill
    \begin{subfigure}[b]{0.45\textwidth}
        \centering
        \includegraphics[width=\linewidth]{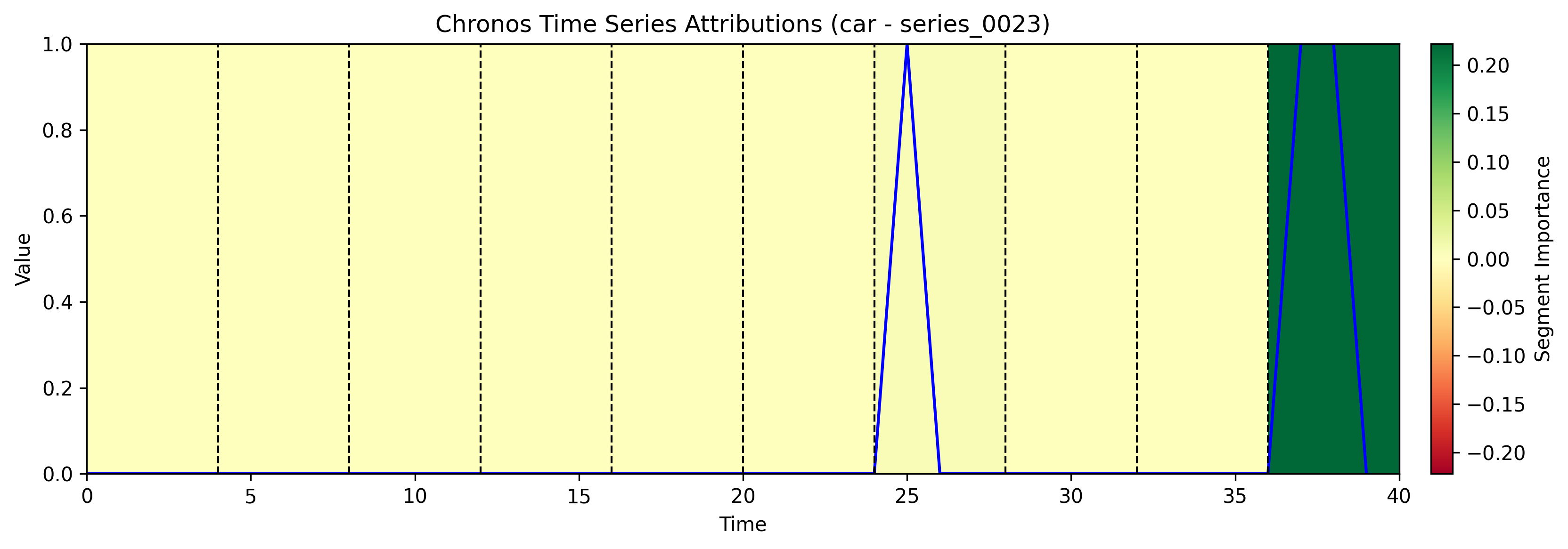}
        \caption{Car series\_23}
        \label{fig:Chronos_LIME_car23_zero}
    \end{subfigure}
    \hfill
    \begin{subfigure}[b]{0.45\textwidth}
        \centering
        \includegraphics[width=\linewidth]{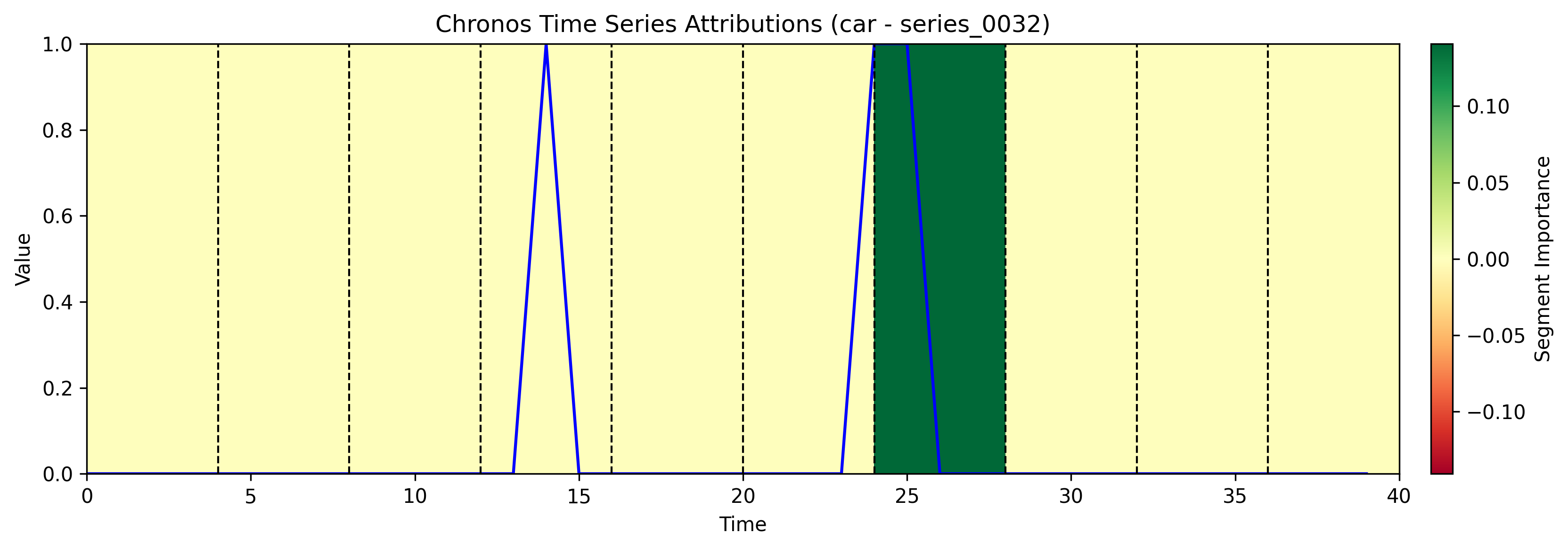}
        \caption{Car series\_32}
        \label{fig:Chronos_LIME_car32_zero}
    \end{subfigure}
    \hfill
    \begin{subfigure}[b]{0.45\textwidth}
        \centering
        \includegraphics[width=\linewidth]{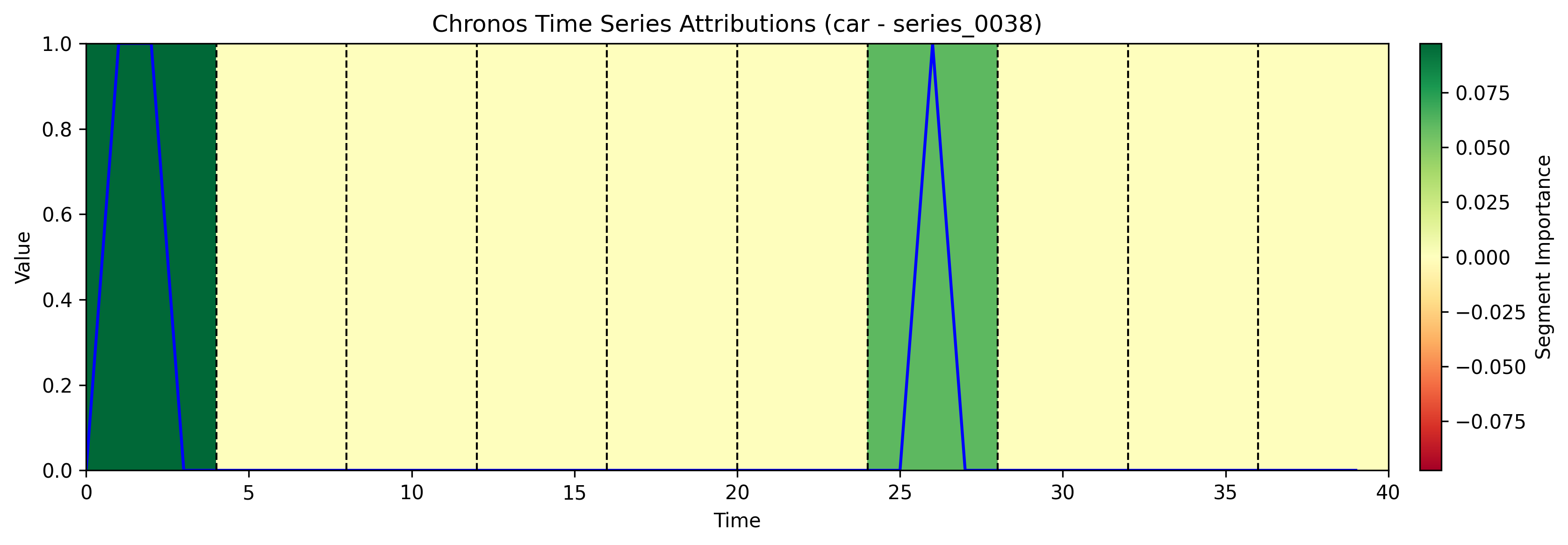}
        \caption{Car series\_38}
        \label{fig:Chronos_LIME_car38_zero}
    \end{subfigure}
    \hfill
    \caption{LIME plots for Chronos model in the car domain using 10 uniform segments, zero replacement, and 100 samples.}
    \label{fig:chronos_lime_values_car_zero}
\end{figure*}

\begin{figure*}[htbp]
    \centering
    \begin{subfigure}[b]{0.45\textwidth}
        \centering
        \includegraphics[width=\linewidth]{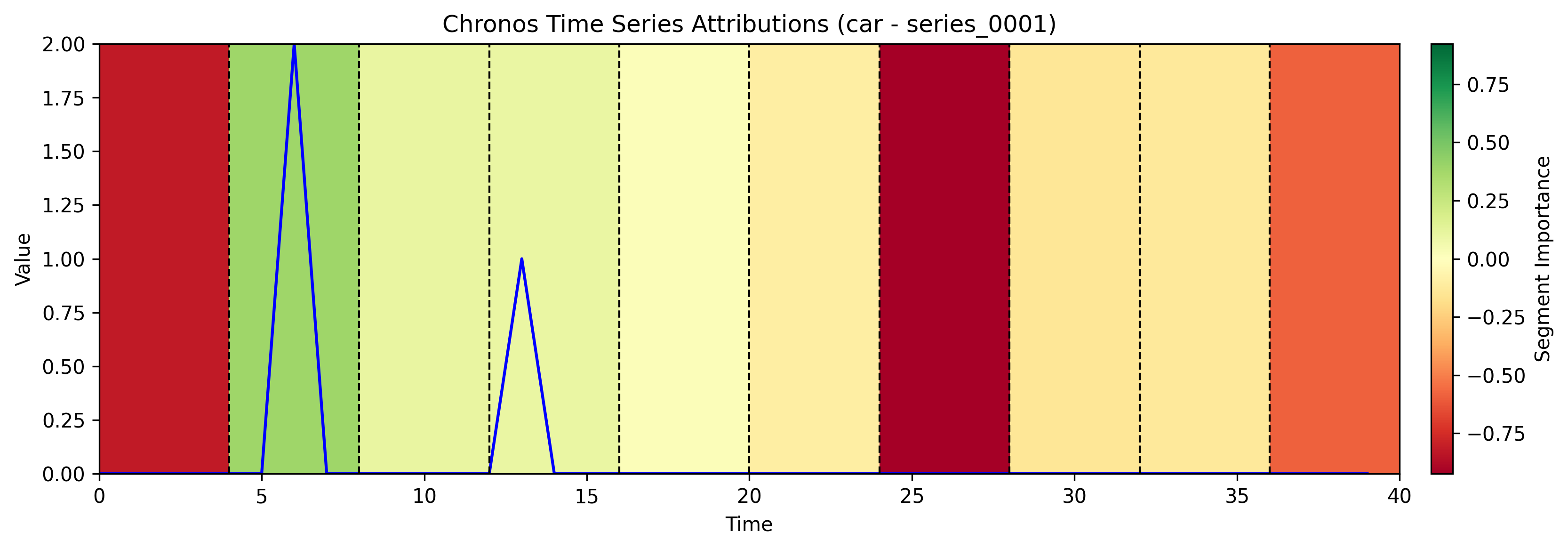}
        \caption{Car series\_1}
        \label{fig:Chronos_LIME_car1_invmax}
    \end{subfigure}
    \hfill
    \begin{subfigure}[b]{0.45\textwidth}
        \centering
        \includegraphics[width=\linewidth]{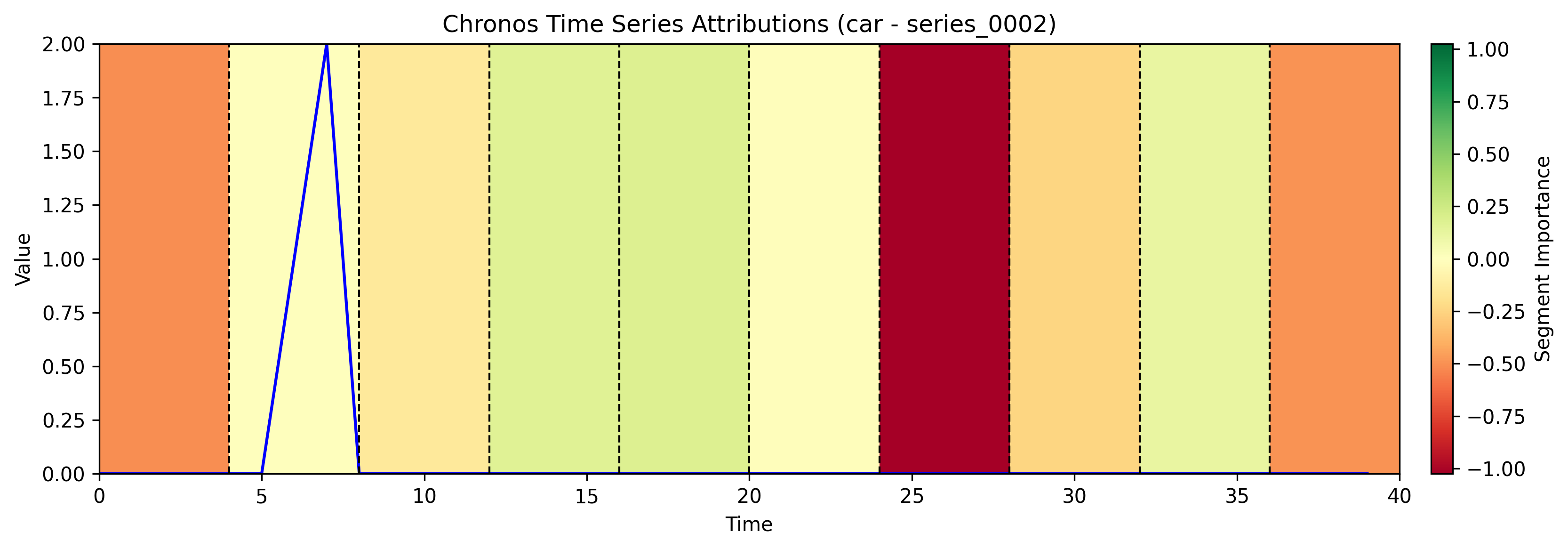}
        \caption{Car series\_2}
        \label{fig:Chronos_LIME_car2_invmax}
    \end{subfigure}
    \hfill
    \begin{subfigure}[b]{0.45\textwidth}
        \centering
        \includegraphics[width=\linewidth]{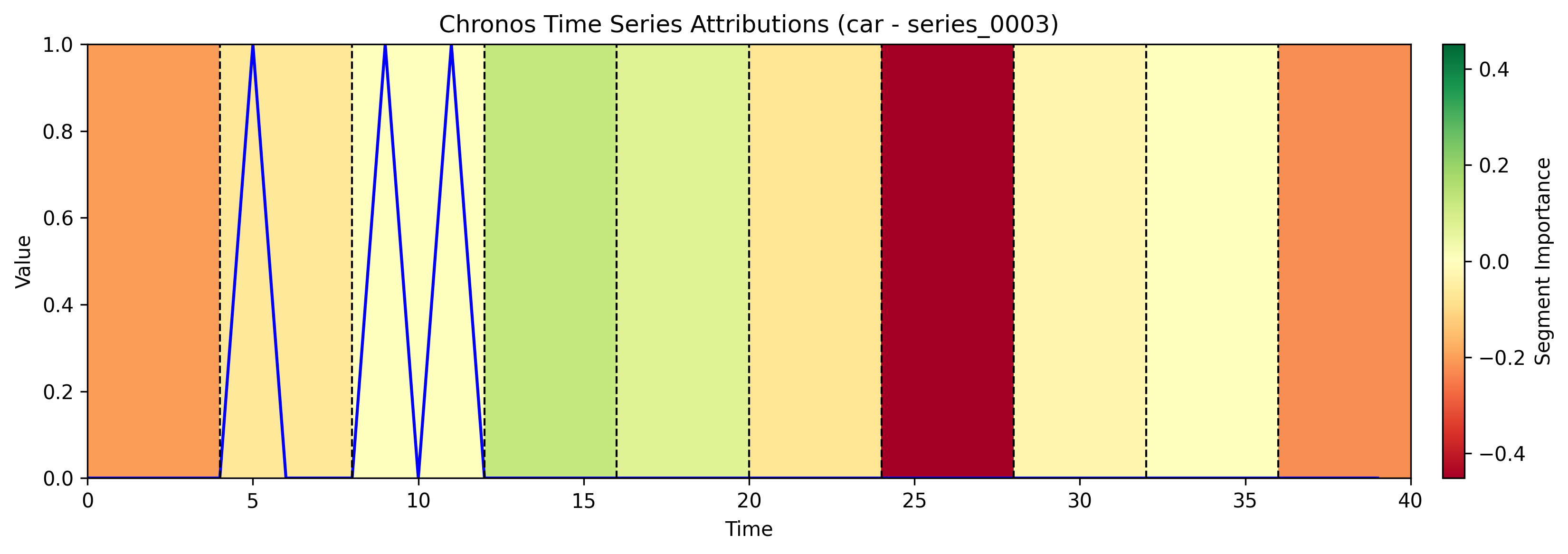}
        \caption{Car series\_3}
        \label{fig:Chronos_LIME_car3_invmax}
    \end{subfigure}
    \hfill
    \begin{subfigure}[b]{0.45\textwidth}
        \centering
        \includegraphics[width=\linewidth]{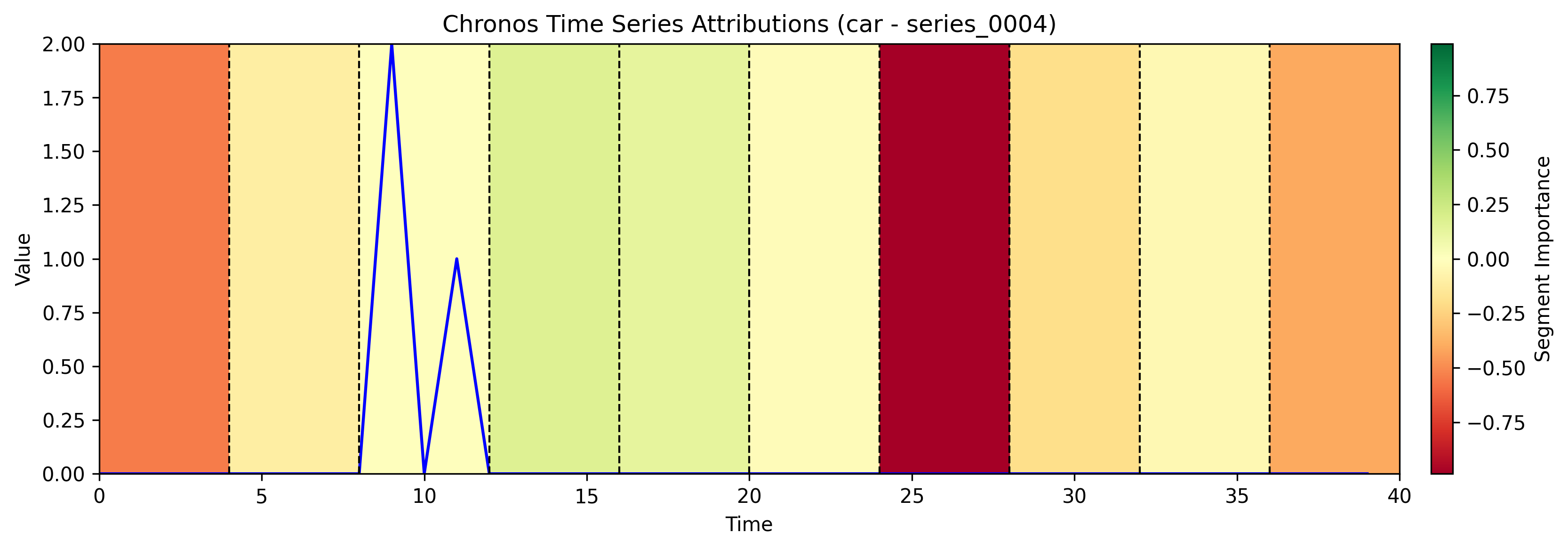}
        \caption{Car series\_4}
        \label{fig:Chronos_LIME_car4_invmax}
    \end{subfigure}
    \hfill
    \begin{subfigure}[b]{0.45\textwidth}
        \centering
        \includegraphics[width=\linewidth]{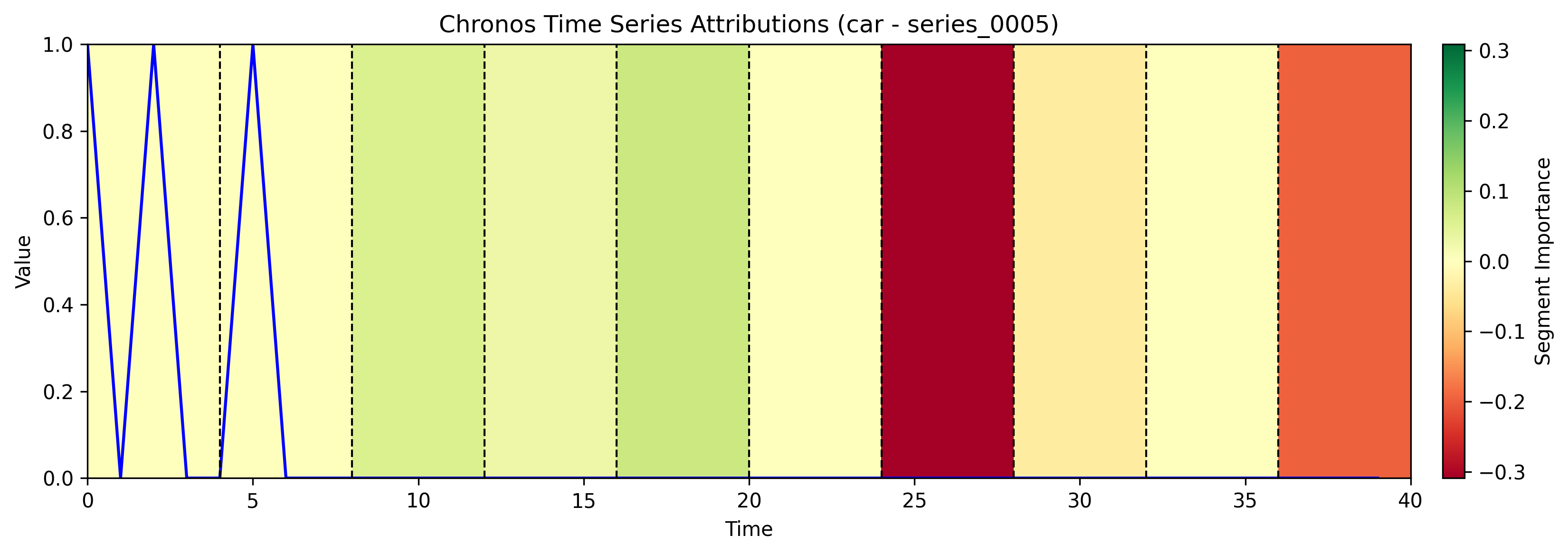}
        \caption{Car series\_5}
        \label{fig:Chronos_LIME_car5_invmax}
    \end{subfigure}
    \hfill
    \begin{subfigure}[b]{0.45\textwidth}
        \centering
        \includegraphics[width=\linewidth]{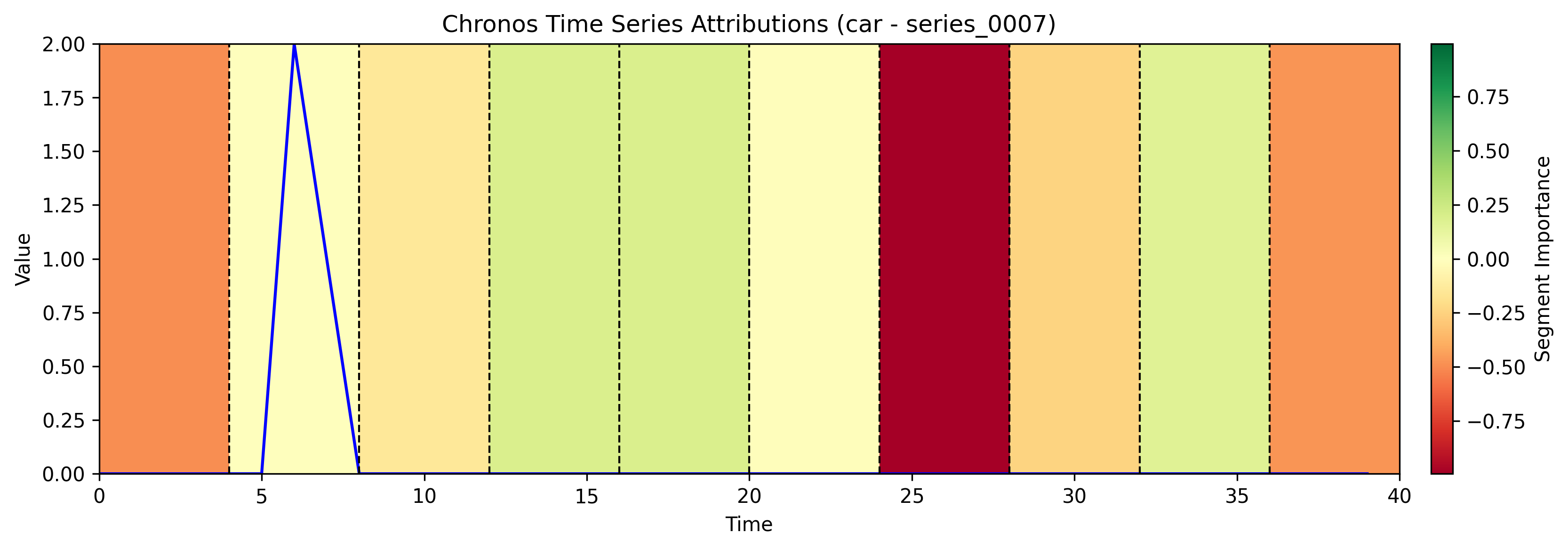}
        \caption{Car series\_7}
        \label{fig:Chronos_LIME_car7_invmax}
    \end{subfigure}
    \hfill
    \caption{LIME plots for Chronos model in the car domain using 10 uniform segments, inverse-max replacement, and 100 samples.}
    \label{fig:chronos_lime_values_car_invmax}
\end{figure*}

\begin{figure*}[htbp]
    \centering
    \includegraphics[width=1.0\textwidth]{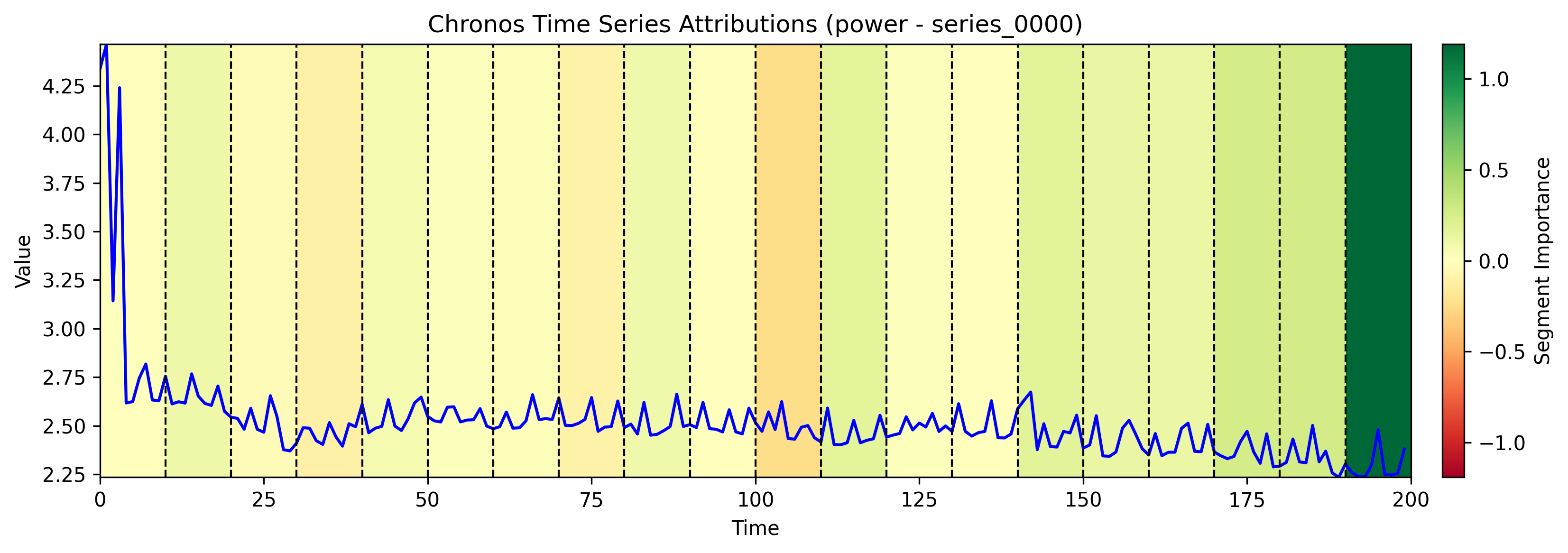}
    \caption{LIME plot for Chronos model in power using 20 uniform segments, zero replacement, and 100 samples on last 200 points of the input time-series.}
    \label{fig:example}
\end{figure*}

Although the Segment LIME explanations provided valuable insights into which portions of the historical data most influenced specific forecasts, they also have notable limitations. The results are highly sensitive to how the time-series is segmented, with different segment lengths or boundaries often producing substantially different importance patterns. The sensitivity to the time-series segmentation can be seen comparing the importances from Figures \ref{fig:chronos_lime_finance_values} and \ref{fig:chronos_lime_values_finance_10seg} which show the method applied to finance with 20 and 10 segments respectively. This sensitivity raises concerns about the stability and reproducibility of the explanations. Furthermore, although the highlighted segments offer an intriguing view of the model’s local dependencies, their direct usefulness for decision-making or for building a clear mental model of the underlying forecasting process remains uncertain. In practice, these explanations should be interpreted as exploratory aids rather than definitive accounts of model behavior.

Similar concerns are raised when looking at the effect of perturbation method. In the car domain, we tested zero perturbation (\ref{fig:chronos_lime_values_car_invmax}) and inverse-max perturbation (\ref{fig:chronos_lime_values_car_zero}), the former replacing targeted segments with zeros and the latter replacing targeted segments with the inverse of the maximum value in the time-series. The car domain is unique in the sparseness of part sales. This property of the data caused most explanations show zero feature importance for all segments when using zero perturbation. This highlights a constraint of perturbing time-series data in that the strategy that is chosen to remove information can be dependent on several factors which include the model used and the time-series data itself.

In pedestrian, figure \ref{fig:chronos_lime_pedestrian_values} shows a similar distribution of feature importance, but has a noticeable bias of negative feature importance for earlier segments and higher feature importance for later ones. The explanation also shows that the majority of segments that fully capture high-traffic days with peaks at rush hours significantly drive the forecast down.
%\xander{@John move the above commented section to appendix with relevant figure}

It should be noted that explanations for ARIMA on the pedestrian and power time-series were omitted due to the infeasibility of generating them. The implementation used in this paper refits the ARIMA model for each step in the forecast, causing it to scale poorly on both input length and forecast horizon, making these explanations infeasible for practical use.

\end{document}